\documentclass[11pt]{article}

\usepackage[preprint]{acl}

\usepackage{times}
\usepackage{latexsym}
\usepackage{lipsum}
\usepackage{fontawesome5}
\usepackage{placeins}

\usepackage[T1]{fontenc}

\usepackage[utf8]{inputenc}

\usepackage{microtype}

\usepackage{inconsolata}

\usepackage{graphicx}
\usepackage{amsmath,amssymb}
\usepackage{tcolorbox}
\usepackage{xcolor}  
\usepackage{colortbl}
\usepackage{booktabs}
\usepackage{enumitem}
\usepackage{fontawesome5}  
\usepackage{xspace}        

\definecolor{myMidBlue}{RGB}{195, 225, 250}
\definecolor{myLightBlueFrame}{RGB}{97, 162, 220} 
\definecolor{mySoftBlue}{RGB}{205, 230, 255} 
\definecolor{myVeryLightBlue}{RGB}{220, 235, 255}   
\definecolor{myExtremelyLightBlue}{RGB}{235, 245, 255}

\definecolor{SystemColor}{HTML}{7E589D}
\definecolor{TableColor}{HTML}{1F77B4}
\definecolor{QuestionColor}{HTML}{FF7F0E}

\newcommand{\systemtag}[1]{\textcolor{SystemColor}{\textbf{#1}}}
\newcommand{\tabletag}[1]{\textcolor{TableColor}{\textbf{#1}}}
\newcommand{\questiontag}[1]{\textcolor{QuestionColor}{\textbf{#1}}}

\newcommand\blfootnote[1]{%
  \begingroup
  \renewcommand\thefootnote{}\footnote{#1}%
  \addtocounter{footnote}{-1}%
  \endgroup
}

\title{A Closer Look into LLMs for Table Understanding}

\author{Jia Wang$^{1,2*}$,\ Chuanyu Qin$^{1,2*}$,\ Mingyu Zheng$^{1,2*}$,\ Qingyi Si$^{3}$, Peize Li$^{1}$, \\ {\bf Zheng Lin$^{1,2\ddagger}$}  \\
$^1$Institute of Information Engineering, Chinese Academy of Sciences, Beijing, China \\
$^2$School of Cyber Security, University of Chinese Academy of Sciences, Beijing, China \\
$^3$JD.COM \\
\texttt{\{wangjia,linzheng\}@iie.ac.cn}\\
}

\begin{document}
\maketitle

\begin{abstract}
Despite the success of Large Language Models (LLMs) in table understanding, their internal mechanisms remain unclear. In this paper, we conduct an empirical study on 16 LLMs, covering general LLMs, specialist tabular LLMs, and Mixture-of-Experts (MoE) models, to explore how LLMs understand tabular data and perform downstream tasks. Our analysis focus on 4 dimensions including the attention dynamics, the effective layer depth, the expert activation, and the impacts of input designs. Key findings include: (1) LLMs follow a three-phase attention pattern---early layers scan the table broadly, middle layers localize relevant cells, and late layers amplify their contributions; (2) tabular tasks require deeper layers than math reasoning to reach stable predictions; (3) MoE models activate table-specific experts in middle layers, with early and late layers sharing general-purpose experts; (4) Chain-of-Thought prompting increases table attention, further enhanced by table-tuning. We hope these findings and insights can facilitate interpretability and future research on table-related tasks.
Our code and analysis pipeline is publicly available at \url{https://github.com/JiaWang2001/closer-look-table-llm}

\blfootnote{$^{*}$ Indicates equal contribution.}
\blfootnote{$^{\ddagger}$ Corresponding author: Zheng Lin.}


\end{abstract}

\section{Introduction}
\label{sec:intro}
Tables, as a representative form of structured data, are widely used across various real-world fields to store and present information. In recent years, Large Language Models (LLMs) have shown powerful instruction following and complex reasoning ability, and thus have become the dominant paradigm for table understanding technique, supporting a wide range of downstream application scenarios, such as table question answering (TQA)~\cite{WTQ,zhang-etal-2024-tablellama,tama}, table fact verification (TFV)~\cite{TabFact,chain_of_table}, advanced data analysis~\cite{CoTA_bench,UniDataBench,ChatGPT_data_analysis} and spreadsheet manipulation~\cite{li2023sheetcopilot,Excel_copilot}.

Although remarkable success has been achieved with existing LLM-based table understanding approaches, they predominantly focus on how to continually improve performance metrics across downstream tasks, e.g., boosting LLMs' performance with the most suitable prompt designs~\cite{sui2024tablemeetsllmlarge,dater_method} and developing specialist tabular LLMs with supervised fine-tuning (SFT)~\cite{tama,zhang-etal-2024-tablellama,zhang-etal-2025-tablellm} or reinforcement learning (RL)~\cite{table_r1_zero,table_r1_teleai,multimodal_table_grpo}. Unlike previous performance-oriented work, in this paper, we conduct an in-depth empirical study to achieve a deeper understanding of the underlying mechanisms of LLMs' table understanding. Our experiments cover a wide spectrum of LLMs of different sizes and types, and focuses on four perspectives including the attention dynamics, the effective layer depth, the impact of MoE architectures and the influence of input designs. The explored research questions are shown in below.



\textbf{Q1:} How do intra-table and extra-table attention patterns of LLMs evolve when performing table understanding tasks?

\textbf{Q2:} How many effective layers are utilized by LLMs to achieve stable predictions and are there difference between tabular and general tasks?

\textbf{Q3:} Are there specialized table experts for MoE models and where are they located?


\textbf{Q4:} The influence of different table formats and reasoning strategies on internal representations.





We conclude our key findings as follows:

\textbf{(1) Different LLM layers exhibit distinct attention patterns during table understanding.} The early layers broadly scan the overall table content, the middle layers then concentrate attention on query-related cells and the top layers further amplify the focused content representations to produce the final answer. 


\textbf{(2) Tabular tasks require more layers for output distribution refinement than math reasoning.} While the depth at which final answer content crystallizes is broadly comparable across tasks, LLMs refine output distributions across more layers when processing tabular data—a pattern consistent across model scales and training strategies.


\textbf{(3) MoE models develop specialized experts for tabular tasks in the middle layers,} which show minimal overlap with math-oriented experts, but early and top layers share general-purpose experts across different tasks.

\textbf{(4) Input formats and reasoning strategies indeed influence attention patterns.} HTML tables result in more dispersed attention distributions in early layers than Markdown tables. Chain-of-Thought (CoT) prompting can allocate more attention to the table content, which can be further amplified by table-specific fine-tuning and thus leads to better performance.


To the best of our knowledge, we present the first thorough investigation of the mechanisms underlying LLM-based table understanding. Our findings not only provide new insights into LLM interpretability in the tabular domain, but also offer guidelines for future research—from optimal input configurations that maximize attention to relevant table content, to inference-time interventions in later layers, to MoE optimizations for table-specialized experts.

\section{Related Work}
\label{sec:related}
\subsection{LLM-based Table Understanding}
Table understanding (TU) technique aims to enable models to automatically comprehend tables and perform various tabular tasks based on user requests~\cite{tu_survey_2023,llm_for_table_processing_survey}. With the rapid iteration of LLMs, their ability has opened new possibilities for more intelligent TU applications. One line of research endeavours to enhance LLMs' table understanding ability through different strategies. For instance, eliciting correct reasoning steps with prompt engineering and carefully selected in-context examples~\cite{table_cot,dater_method,chain_of_table,Tree_of_Table,jiang-etal-2023-structgpt}, collecting table instruction tuning data for supervised fine-tuning~\cite{zhang-etal-2024-tablellama,zhang-etal-2025-tablellm}, exploring new tasks and reward strategies for reinforcement learning~\cite{table_r1_zero,table_r1_teleai,p2_tqa}, and building powerful table agents for multi-turn data analysis and excel 
manipulation~\cite{li2023sheetcopilot,UniDataBench}.

In addition to the performance-oriented studies mentioned above, another valuable direction is to explore the robustness and interpretability of LLMs' table-related capabilities. For example, evaluating models' performance against divergent perturbations such as table formats~\cite{sui2024tablemeetsllmlarge}, noisy operators~\cite{tabular_repr_impacts_on_llm_tu} and cell replacement~\cite{explore_robust_of_llm_tu_via_attention_analysis}. Compared with existing work that mainly focused on the performance robustness, we aim at seeking a better understanding of the underlying mechanisms of LLM-based table understanding and provide valuable insights for future follow-ups.

\subsection{Interpretability and Analysis of LLMs}


Despite the strong capabilities of LLMs, understanding their internal mechanisms remains a key challenge. Some studies focus on the roles of different layers in LLMs, examining how individual layers contribute to information processing~\cite{skean2025layer} and how semantic representations are gradually constructed across layers~\cite{csordas2025_effective_depth, hu2025affects_effective_depth}. Other lines of work investigate specific internal components, such as attention heads in self-attention mechanisms and feed-forward network (FFN) modules, with the aim of clarifying their roles in knowledge representation~\cite{geva2021transformer, meng2022locating} and information acquisition~\cite{wu2024retrieval_head, kobayashi2023analyzing}, revealing various intriguing phenomena within LLMs, such as attention sink~\cite{kobayashi-etal-2020-attention, xiao2023streamingllm, gu2024attention, qiu2025gatedattentionlargelanguage, queipo2025attention}. Furthermore, with the growing adoption of Mixture-of-Experts (MoE) architectures, recent researches~\cite{ESFT, routertune, lo2025closer, bandarkar2025multilingual, MoE_Super_Expert} analyze expert modules within MoE models, exploring their specialization patterns and activation behaviors during task execution.

While these efforts have provided valuable insights~\cite{si-etal-2023-alpaca_cot,llm_judge_empirical_study}, they predominantly focus on unstructured text tasks such as math reasoning and factual recall. In this work, we extend interpretability research to structured table data, conducting analyses to explore how LLMs process tables and offering preliminary findings across several dimensions.


\section{Experimental Setups}
\label{sec:setups}
\paragraph{Evaluation Data}
\label{sec:eval_data}
We randomly select 500 samples from 3 TQA benchmarks and 1 TFV benchmark to perform empirical study, which include WTQ~\citep{WTQ}, HiTab~\citep{Hitab}, AIT-QA~\citep{AITQA} and TabFact~\citep{TabFact}. The resulting 2,000 test samples cover common table structures with flat headers (WTQ, TabFact) and complex hierarchical headers (HiTab, AIT-QA). The input tables are serialized into Markdown format by default, and we also analyze the impact of alternative formats (e.g., HTML) in Section~\ref{sec:rq4_format}. The complete input prompt templates are shown in Appendix~\ref{prompt_template}.

\begin{table}[t]
\centering
\caption{Summary of 2,000 evaluation samples used in our empirical study.}
\label{analysis_datasets_info}
\small
\begin{tabular}{@{}l l l c c@{}}
\toprule
\textbf{Dataset} & \textbf{Task} & \textbf{Structure} & \textbf{Avg Tokens} & \textbf{\#Samples} \\
\midrule
WTQ      & TQA & Flat         & 1012 & 500 \\
HiTab    & TQA & Hierarchical & 964  & 500 \\
AITQA    & TQA & Hierarchical & 651  & 500 \\
TabFact  & TFV & Flat         & 681  & 500 \\
\bottomrule
\end{tabular}
\end{table}

\paragraph{Analyzed LLMs}
\label{sec:analyzed_llms}
We analyze 16 LLMs of three categories. (1) general-purpose instruct LLMs such as Llama-3.1-8B-Instruct and Qwen-2.5-7B-Instruct. (2) fine-tuned tabular LLMs via SFT or RL such as TAMA~\citep{tama} and Table-R1-Zero~\citep{table_r1_zero}. (3) Mixture-of-Experts (MoE) models like DeepSeek-V2-Lite~\citep{deepseek_v2} and Qwen3-30B-A3B~\citep{qwen3}. The complete model list is provided in Appendix~\ref{appendix::used_llms}.

\section{Analysis of Attention Dynamics}
\label{sec:rq1}

To understand how LLMs allocate attention during table question answering, we design a controlled experimental setup where each input instance comprises three segments: (1) a \systemtag{system prompt} providing general instructions, (2) a \tabletag{table content} containing the serialized table, and (3) a \questiontag{user question} specifying the query. This segmentation allows us to trace attention flow across different input components. We investigate three progressive questions: \textbf{\S\ref{sec:rq1_attention}}: How does the model attend to three input segments, especially different table cells? \textbf{\S\ref{sec:rq1_contribution}}: How much does the table content actually contribute to the model's final output? \textbf{\S\ref{sec:rq1_validation}}: Does this attention pattern causally affect model predictions?

The following metrics are employed in our analysis. (1) \textit{segment attention ratio} measures the proportion of attention allocated to each segment per layer; (2) \textit{table attention entropy} quantifies how concentrated the attention is within the table (lower entropy indicates a more focused attention distribution on specific cells); and (3) \textit{attention contribution} captures the actual influence of each segment on the final output via L2 norm of value-weighted representations. Formal definitions and formulas of these metrics are provided in Appendix~\ref{sec:methods}. For all metrics, we first average across all generated tokens of each sample, and then average across all 2,000 test samples. The attention dynamics of Llama3.1-8B-Instruct and Qwen2.5-7B-Instruct is shown in Figure~\ref{main_analysis_pic}. Results of more models are shown in  Appendix~\ref{sec:appendix_attention_all}.


\begin{figure*}[!t]
  \centering
  \includegraphics[width=\linewidth]{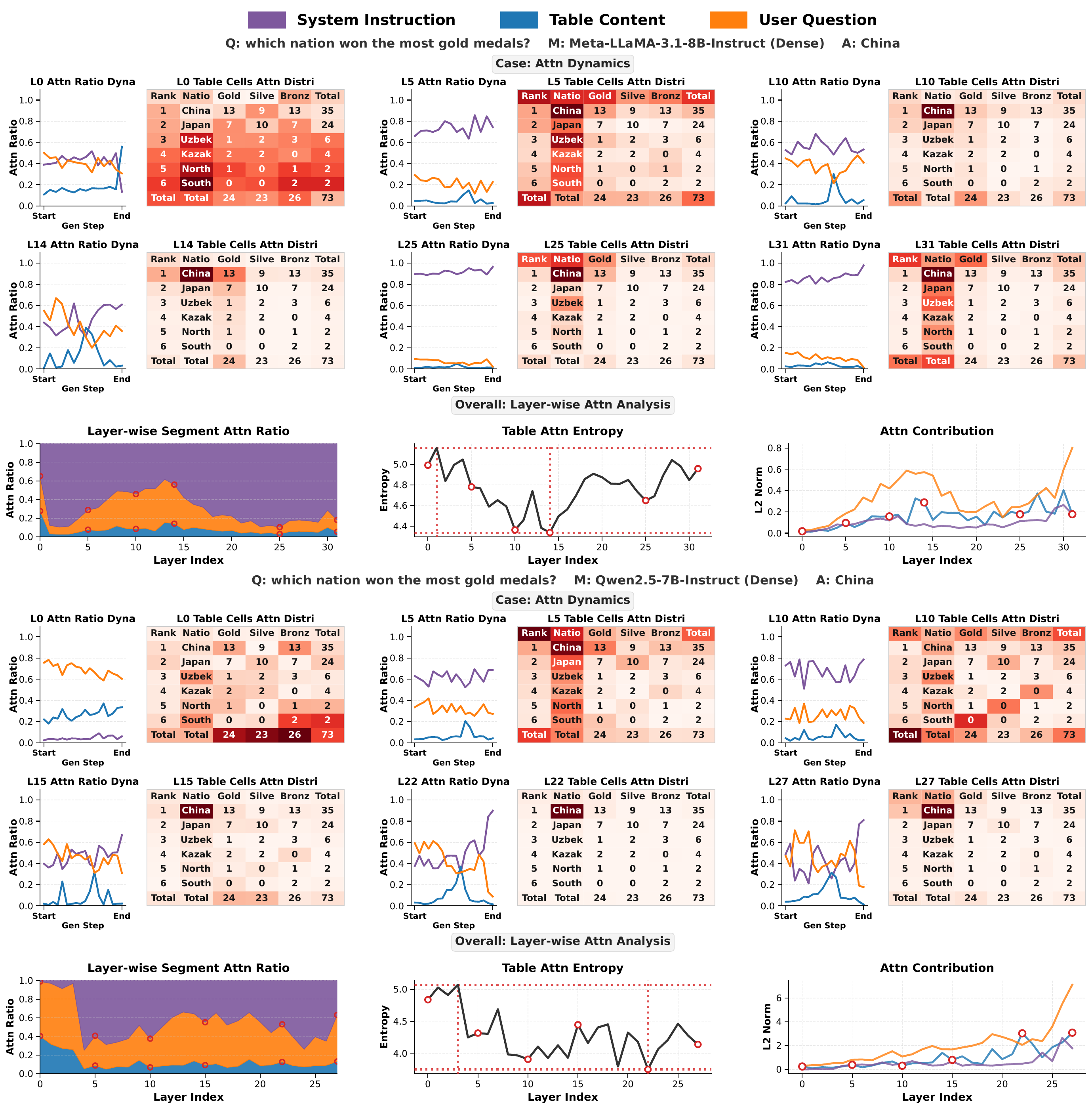}
  \caption{The attention dynamics of Llama3.1-8B-Instruct and Qwen2.5-7B-Instruct on tabular tasks. Each input consists of three segments: \systemtag{system prompt}, \tabletag{table content}, and \questiontag{user question}. \textbf{Upper}: A case study at selected layers---left sub-panels show step-wise attention ratio trend throughout generation steps; right sub-panels show cell-level attention heatmaps. \textbf{Lower}: Aggregated results---\textit{Layer-wise Segment Attn Ratio} shows the proportion of attention allocated to each segment per layer; \textit{Table Attn Entropy} measures the degree of focus towards table cells, with lower entropy indicating a more concentrated attention distribution on specific cells); \textit{Attn Contribution} measures the influence of each segment on the  final output. Notably, the entropy minimum at Layer 10 and Layer 14 for Llama3.1-8B-Instruct align with the concentrated attention on the answer cell ``China'' in the upper heatmap.}
  \label{main_analysis_pic}
\end{figure*}

\subsection{How Does the Model Attend to the Table?}
\label{sec:rq1_attention}

\begin{tcolorbox}[colback=cyan!5!white, colframe=cyan!45!blue!60, title=\textbf{Takeaway 1}]
Although the overall attention ratio to table content is modest compared to other segments, LLMs can precisely focus on query-relevant cells in the middle layers, suggesting that accurate localization matters more than total attention volume.
\end{tcolorbox}

\paragraph{Allocated attention ratio across input segments.} 


As shown in Figure~\ref{main_analysis_pic} (lower left), the attention allocated to \tabletag{<table content>} is highest in the earliest layers, then it drops before rising again to a secondary peak in the lower-middle layers (layer 10-15) and finally decreases to a stable level in the top layers (layer 20-30). Notably, the overall changes of table-oriented attention across different layers are modest. We observe a similar trend of attention allocated to \questiontag{<user question>}, with both reaching their peak in the middle layers, which may suggest that the model is integrating question information to interpret table structures and identify critical table content. Except for middle layers, the \systemtag{<system prompt>} consistently attracts relatively higher attention scores across all layers than user question and table content, which could be attributed to the attention sink phenomenon observed in prior work~\citep{xiao2023streamingllm, gu2024attention, kang2025see}. The step-wise segment attention ratio in specific layers throughout the token-by-token generation process of one sample is also visualized in the upper left sub-panels in Figure~\ref{main_analysis_pic}, which provides evidence of observed patterns.

\paragraph{The influence of model scales.} Comparing models of different scales within the same family (detailed results in Appendix~\ref{sec:appendix_attention_all}), we find that overall attention distribution trends remain consistent. However, two notable differences emerge: (1) the secondary peak of table attention shifts toward deeper layers as model size increases, and (2) the layer range with low table attention entropy becomes broader in larger models, suggesting they maintain focused attention across more layers.

\paragraph{The influence of model architectures.}
Comparing Llama3.1-8B and Qwen2.5-7B (Figure~\ref{main_analysis_pic}), we observe that Qwen allocates higher attention to table content in early layers. MoE models exhibit a more gradual decline in table attention entropy, with a distinctive three-phase pattern: attention first covers partial cells, then spreads across the entire table, and finally converges to answer-relevant cells (detailed in \S\ref{sec:rq3}). For tabular LLMs like TableGPT2~\cite{tablegpt2}, table-specific fine-tuning leads to higher table attention and weaker attention sink effects, though overall patterns remain similar to their base models.


\paragraph{Attention distribution within the table.}
While the overall attention allocated to table is limited, does the model at least focus on the query-related table cells? To answer this, we measure the \textit{table attention entropy} over table tokens at each layer. 


As shown in Figure~\ref{main_analysis_pic} (lower middle), the table attention entropy follows a U-shaped pattern: it decreases from early layers, and reaches a minimum in the middle layers, and finally increases again. The attention heatmaps in Figure~\ref{main_analysis_pic} (upper right sub-panels) also illustrate this pattern. In early layers (e.g., Layer 0), attention is spread broadly across the entire table, suggesting that the model scans all content. In middle layers (e.g., Layer 14 for LLaMA-3.1-8B-Instruct), attention becomes concentrated on specific cells---for the question ``which nation won the most gold medals?'', the model focuses on the ``China'' cell and the ``Gold'' column header, which directly correspond to the answer. In later layers, attention spreads slightly again as the model prepares to generate the answer. This pattern is consistent across different models and benchmarks. We provide more case studies and model comparisons in the Appendix~\ref{sec:appendix_attention_all}.

These findings suggest that LLMs do not require heavy attention to all table content. Instead, they selectively focus on query-relevant cells, much like humans skimming a table for key information. 
This implies that low-attention table regions could be pruned to develop efficient tabular LLMs with less performance loss.

\subsection{How Much Does the Table Content Contribute to the Final Output?}
\label{sec:rq1_contribution}

\begin{tcolorbox}[colback=cyan!5!white, colframe=cyan!45!blue!60, title=\textbf{Takeaway 2}]
Despite receiving modest attention, table content makes substantial contributions to the model's output, with its influence increasing in later layers and ultimately driving answer generation.
\end{tcolorbox}

Attention weights show where the model "looks", but not how much the attended content actually influences the output. As noted by prior work~\citep{kobayashi-etal-2020-attention, gu2024attention}, the true impact depends on both the attention weight and the transformed value vector. Therefore, we measure the \textit{value-weighted contribution}---the L2 norm of each segment's representation injected into the transformer residual stream (see Appendix~\ref{sec:methods} for formal definition).

As shown in Figure~\ref{main_analysis_pic} (lower right), despite the modest attention ratio observed in \S\ref{sec:rq1_attention}, the table segment makes substantial contributions that increase in later layers, peaking near the final layers. In contrast, the system prompt's contribution remains low across all layers despite its high attention ratio, confirming its role as an attention sink rather than an information source.

This finding connects previous observations into a coherent narrative of the model's internal workflow. While early layers broadly encode the table (high ratio, high entropy), middle layers act as a critical reasoning phase characterized by "concentrated influence": the model narrows focus to relevant cells (low entropy) and amplifies their impact on the residual stream (rising norm). Finally, in late layers, the question and focused table content jointly drive answer generation.

\subsection{Does the Attention Pattern Causally Affect Output?}
\label{sec:rq1_validation}

\begin{tcolorbox}[colback=cyan!5!white, colframe=cyan!45!blue!60, title=\textbf{Takeaway 3}]
The three-phase attention pattern is functionally essential rather than a mere processing correlate: masking table attention in early-to-middle layers causes severe performance degradation, and middle-layer attention focus on answer cells is predictive of output correctness.
\end{tcolorbox}

\begin{figure}[t]
  \centering
  \includegraphics[width=\linewidth]{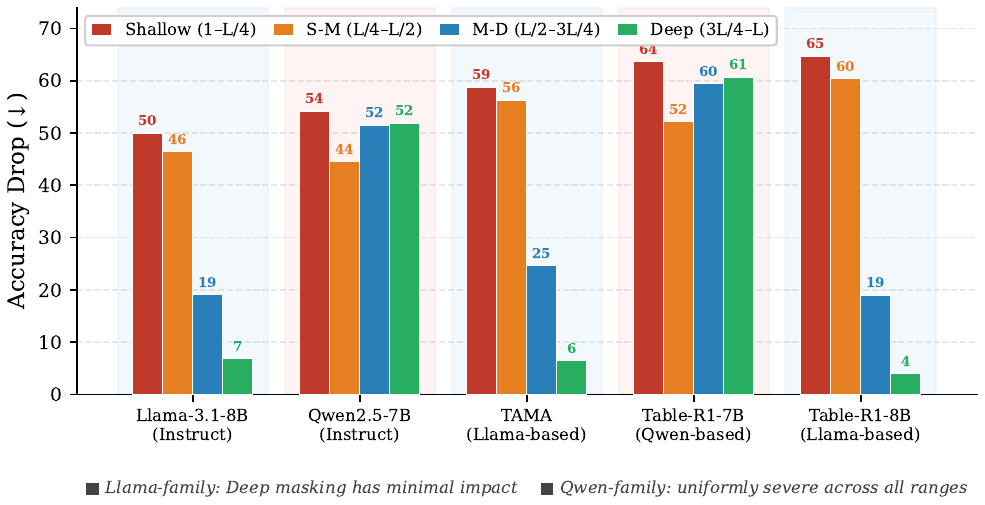}
  \caption{Performance drop when masking table attention in different layer ranges across 5 models (averaged over 3 benchmarks $\times$ 2 formats). Llama-family models show minimal impact from Deep masking, while Qwen-family models degrade uniformly across all ranges.}
  \label{fig:masking_results}
\end{figure}

\paragraph{Causal validation via attention masking.}
We zero out attention weights to all table tokens within consecutive quarter-layer blocks---Shallow (1--$L$/4), S-M ($L$/4--$L$/2), M-D ($L$/2--3$L$/4), and Deep (3$L$/4--$L$)---across 5 models, 3 benchmarks, and 2 formats. As shown in Figure~\ref{fig:masking_results}, masking early layers (Shallow, S-M) universally causes the largest performance drops (44--65 points), directly demonstrating that the broad scanning phase is functionally essential to downstream prediction. Llama-family models (Llama-3.1-8B, TAMA, Table-R1-8B) show the most severe degradation from Shallow and S-M masking with minimal late-layer impact (TAMA drops by 58.8 and 56.3 points for Shallow and S-M, respectively, but only 6.5 points for Deep), empirically validating our interpretation that late layers serve an amplification rather than retrieval role. In contrast, Qwen-family models (Qwen2.5-7B, Table-R1-7B) show uniformly severe degradation across all masking ranges, reflecting their more distributed attention profiles (\S\ref{sec:rq1_attention}).

\paragraph{Attention-answer correspondence.}
To complement the causal analysis, we select 200 lookup-type examples and identify the Top-20 attention-weighted cells in low-entropy layers (layers 10--15) for Llama-3.1-8B and Qwen2.5-7B. Cells focused on by low-entropy layers directly correspond to the model's final response in 53.0\% (Llama) and 43.9\% (Qwen) of cases. When low-entropy layers successfully attend to the correct answer cell, final prediction accuracy reaches 65.6\% (Llama) and 71.2\% (Qwen). Conversely, when models answer incorrectly, 52.8\% (Llama) and 74.4\% (Qwen) of errors are associated with low-entropy layers failing to attend to the correct cell. This suggests that misfocused middle-layer attention is a frequent predictor of incorrect predictions, pointing to potential utility for attention-based error detection.

\section{Effective Depth for Tabular Tasks}
\label{sec:rq2}

The previous analysis revealed that middle layers concentrate attention on query-relevant cells while later layers amplify their contributions. A natural follow-up question is: how many layers do LLMs actually need to complete table understanding tasks? To investigate this, we apply LogitLens~\citep{logitlen}, which decodes hidden representations at each layer into vocabulary distributions. We measure the KL divergence between each layer's prediction and the final output, along with the top-5 token overlap.

\begin{figure}[t]
  \centering
  \includegraphics[width=\linewidth]{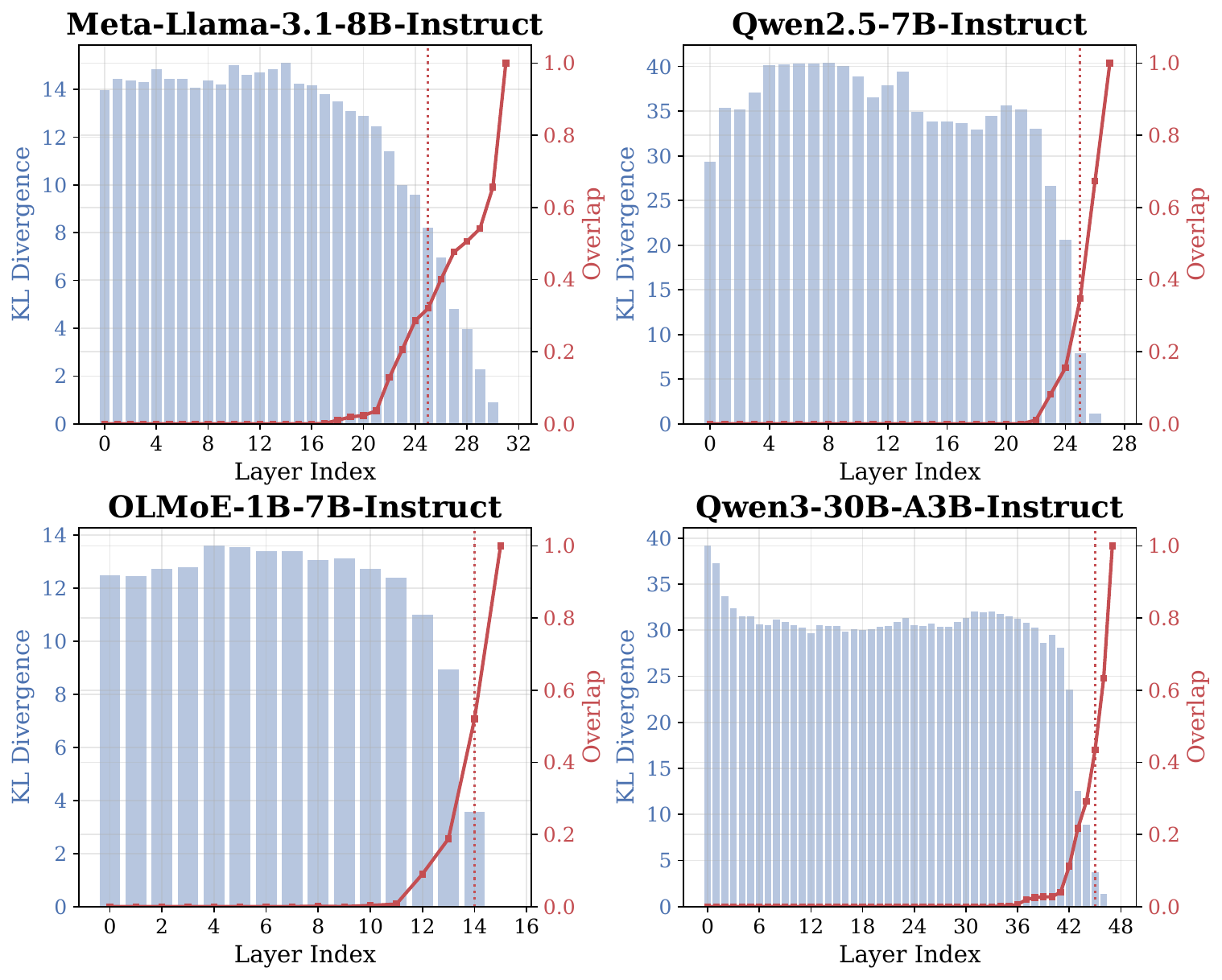}
  \caption{Prediction stability analysis using LogitLens. Bars show KL divergence between each layer's decoded distribution and the final output (lower = closer to final prediction). Lines show top-5 token overlap with the final output. Vertical dashed lines mark where predictions stabilize.}
  \label{effective_depth_analysis}
\end{figure}

\begin{tcolorbox}[colback=cyan!5!white, colframe=cyan!45!blue!60, title=\textbf{Takeaway 3}]
Table tasks engage more layers for output distribution refinement than math reasoning, though the depth at which answer content crystallizes is comparable across tasks. This pattern is consistent across model scales and fine-tuning strategies, suggesting that layer-wise functionality is largely determined during pre-training.
\end{tcolorbox}

As shown in Figure~\ref{effective_depth_analysis}, KL divergence remains relatively high throughout the middle layers and only drops sharply in final layers. To directly validate whether this pattern distinguishes table tasks from other reasoning tasks, we run the same LogitLens pipeline on GSM8K across 7 representative models spanning general instruction-tuned, table-specific fine-tuned, and MoE categories. We decompose stabilization depth into two complementary metrics: \textit{KL-based depth} (where output distribution converges) and \textit{Top-$k$ overlap depth} (where the final answer content crystallizes).

\begin{table}[t]
\centering
\caption{Stabilization depth comparison between table and math (GSM8K) tasks. Values are reported as effective layer / total layers (ratio). KL-based depth is higher for table tasks in 5 of 7 models ($\dagger$), while Top-$k$ overlap depth is broadly comparable.}
\resizebox{\columnwidth}{!}{
\begin{tabular}{lc cc cc}
\toprule
 & & \multicolumn{2}{c}{\textbf{KL Depth}} & \multicolumn{2}{c}{\textbf{Overlap Depth}} \\
\cmidrule(lr){3-4} \cmidrule(lr){5-6}
\textbf{Model} & \textbf{Layers} & Table & Math & Table & Math \\
\midrule
\multicolumn{6}{l}{\cellcolor[rgb]{0.957,0.957,0.957}\textit{\textbf{Instruct LLMs}}} \\
\addlinespace[0.3em]
Qwen2.5-3B   & 36 & 32 (0.89)$^\dagger$ & 27 (0.75) & 33 (0.92) & 31 (0.86) \\
Qwen2.5-7B   & 28 & 25 (0.89)$^\dagger$ & 23 (0.82) & 25 (0.89) & 25 (0.89) \\
Qwen2.5-14B  & 48 & 44 (0.92)$^\dagger$ & 43 (0.90) & 44 (0.92) & 43 (0.90) \\
Llama-3.1-8B & 32 & 26 (0.81) & 29 (0.91) & 25 (0.78) & 29 (0.91) \\
\midrule
\multicolumn{6}{l}{\cellcolor[rgb]{0.957,0.957,0.957}\textit{\textbf{Table-Specific Fine-tuned LLMs}}} \\
\addlinespace[0.3em]
TAMA         & 32 & 26 (0.81)$^\dagger$ & 23 (0.72) & 25 (0.78) & 26 (0.81) \\
Table-R1-7B  & 28 & 25 (0.89)$^\dagger$ & 24 (0.86) & 25 (0.89) & 25 (0.89) \\
\midrule
\multicolumn{6}{l}{\cellcolor[rgb]{0.957,0.957,0.957}\textit{\textbf{MoE LLMs}}} \\
\addlinespace[0.3em]
Qwen3-30B-A3B & 48 & 43 (0.90) & 43 (0.90) & 45 (0.94) & 43 (0.90) \\
\bottomrule
\end{tabular}}
\label{tab:gsm8k_comparison}
\end{table}

As shown in Table~\ref{tab:gsm8k_comparison}, KL-based stabilization depth is consistently higher for table tasks than math across the majority of models (5 of 7, marked with $\dagger$), indicating that LLMs continue refining their output distributions across more layers when processing tabular data. Top-$k$ overlap depth, by contrast, is broadly comparable across both task types, suggesting that while the distribution refinement process engages more layers for tables, the point at which final answer content crystallizes is similar. This decomposition provides a more precise characterization: table understanding does not simply ``take longer'' uniformly, but specifically requires more layers for the fine-grained distribution adjustments that follow answer identification.

Notably, this pattern holds consistently across different model scales (7B to 32B) and training strategies (SFT and RL), suggesting that the layer-wise functionality for table understanding is largely established during pre-training (detailed results in Appendix~\ref{sec:appendix_effective_depth}). This finding has practical implications: unlike math reasoning where early-exit strategies can reduce computation~\citep{csordas2025_effective_depth, hu2025affects_effective_depth}, table tasks may benefit less from such optimizations. Conversely, it suggests that targeted interventions in the later layers (e.g., inference-time steering) could be particularly effective for improving table understanding capabilities.

\begin{figure*}[t]
  \centering
  \includegraphics[width=0.85\linewidth]{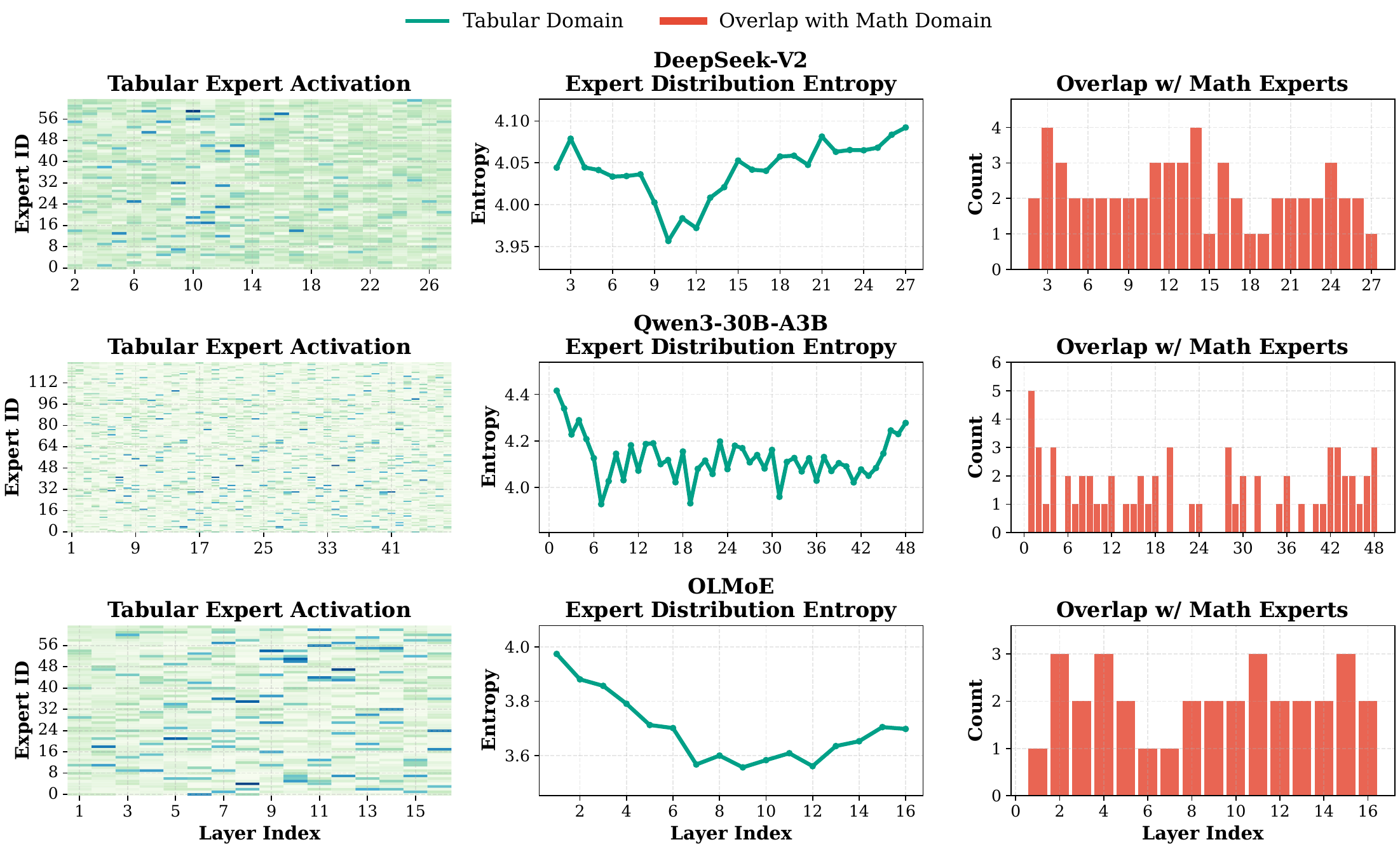}
  \caption{Expert activation analysis in MoE models across three architectures (DeepSeek-V2, Qwen3-30B-A3B, OLMoE). \textbf{Left}: Activation heatmaps showing table-specific experts per layer (darker = higher activation probability). \textbf{Middle}: Entropy of expert activation distribution for table tasks (lower entropy indicates concentrated activation on fewer experts). \textbf{Right}: Number of overlapping experts between table and math (GSM8K) tasks per layer.}
  \label{moe_expert_analysis}
\end{figure*}


\section{Seeking Table Experts of MoE Models}
\label{sec:rq3}

\begin{tcolorbox}[colback=cyan!5!white, colframe=cyan!45!blue!60, title=\textbf{Takeaway 4}]
MoE models activate a distinct set of table-specific experts concentrated in the middle layers, mirroring the attention patterns observed in RQ1. These experts show minimal overlap with math-related experts.
\end{tcolorbox}

Mixture-of-Experts (MoE) models route each token to a subset of experts based on learned gating functions. A natural question arises: do MoE models develop specialized experts for tabular tasks?

To identify table-specific experts, we record the activation frequency of each expert across all table tokens and layers, following prior work~\citep{ESFT, routertune}. As shown in Figure~\ref{moe_expert_analysis} (left), table-relevant experts are distributed across all layers, but their activation patterns vary by depth. The middle column of Figure~\ref{moe_expert_analysis} reveals a striking pattern: the entropy of expert activation distribution reaches its minimum in the middle layers across all three MoE architectures. Lower entropy indicates that the model concentrates its routing on fewer, more specialized experts. This pattern mirrors the attention entropy findings in RQ1, where attention within the table also becomes most concentrated in the middle layers. Furthermore, these table-specific experts in the middle layers show minimal overlap with experts activated for math reasoning tasks (Figure~\ref{moe_expert_analysis}, right), confirming their domain-specific nature.

Together, these findings suggest that MoE models achieve efficient multi-task processing by sharing general-purpose experts in early and late layers while routing to specialized experts in the middle layers for domain-specific reasoning.


\section{Impact of Input Formats and Reasoning Strategies}
\label{sec:rq4}


The previous sections analyzed LLMs' internal mechanisms under a fixed setting (Markdown format, direct answering). In practice, however, tables can be represented in various formats (e.g., Markdown, HTML), and models can be prompted with different reasoning strategies (e.g., direct answering, Chain-of-Thought). Here we examine how these choices influence internal processing and model behavior.

\subsection{How Does Table Format Affect Attention?}
\label{sec:rq4_format}

\begin{figure*}[t]
  \centering
  \includegraphics[width=\linewidth]{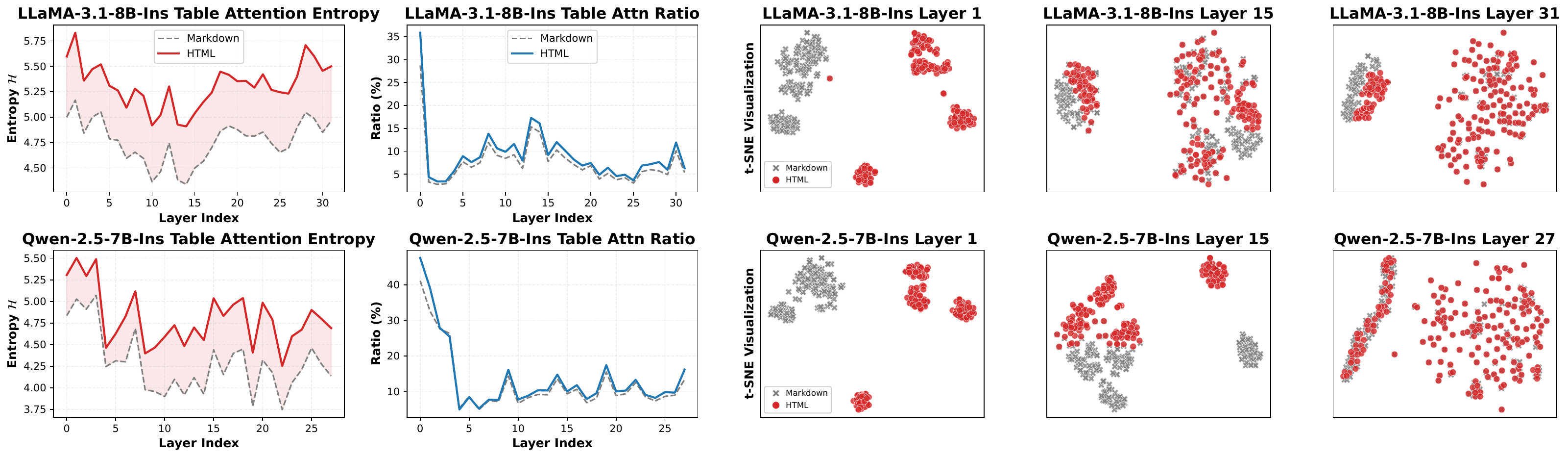}
  \caption{Comparison between Markdown and HTML table formats.}
  \label{table_format_analysis}
\end{figure*}

\begin{figure*}[t]
  \centering
  \includegraphics[width=\linewidth]{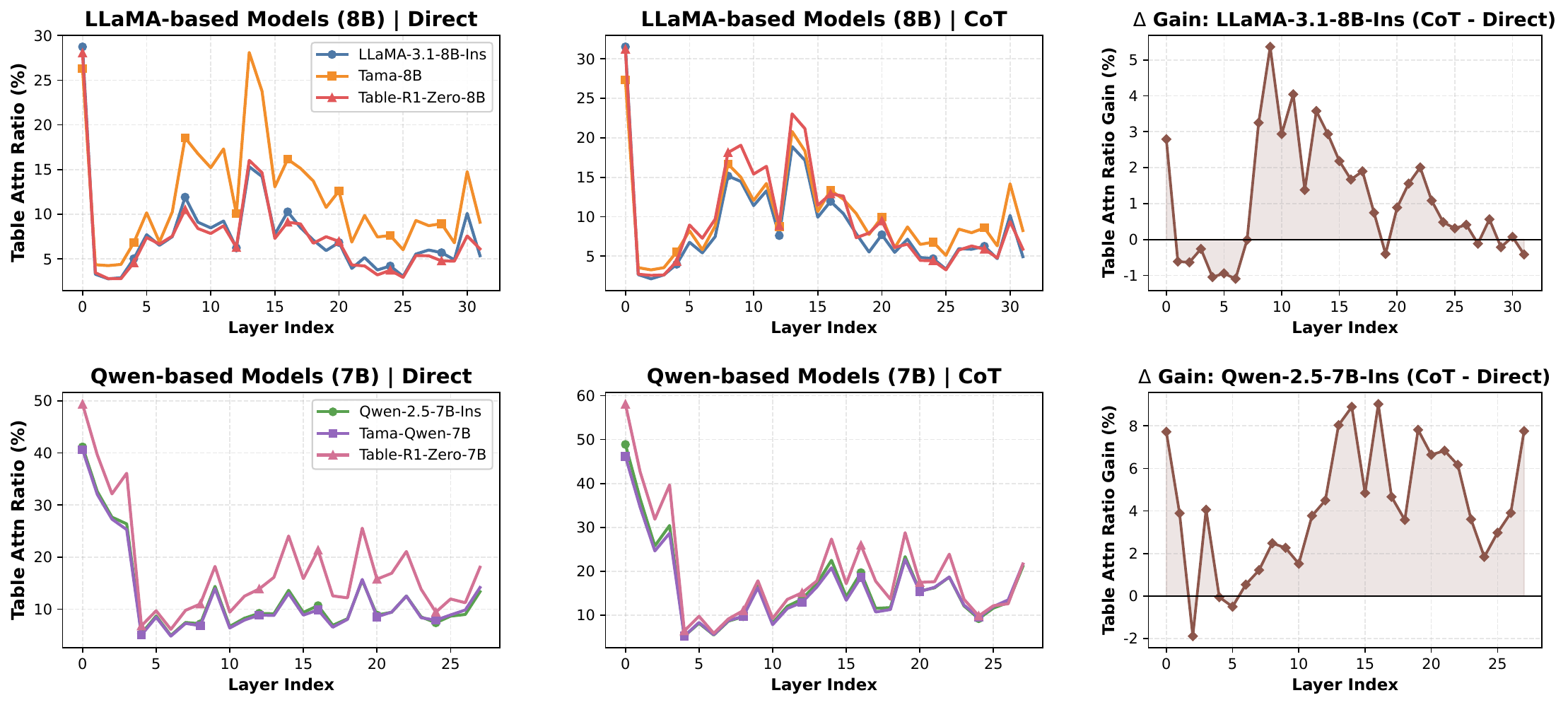}
  \caption{Comparison between direct answering and Chain-of-Thought (CoT) reasoning.}
  \label{reasoning_settings_analysis}
\end{figure*}

\begin{tcolorbox}[colback=cyan!5!white, colframe=cyan!45!blue!60, title=\textbf{Takeaway 5}]
HTML tables lead to higher attention entropy than Markdown tables. However, internal representations of different formats converge in deeper layers.
\end{tcolorbox}

To investigate how table format affects internal processing, we convert our analysis data from Markdown to HTML format and compare the attention patterns. As shown in Figure~\ref{table_format_analysis}, LLMs assign slightly higher attention ratio to HTML tables, but the attention entropy is significantly higher. This may be because the verbose structural syntax in HTML (e.g., \texttt{<tr>}, \texttt{<td>} tags) disperses the model's attention across more tokens, making it harder to concentrate on specific cells.

Despite these differences in early layers, $t$-SNE visualizations reveal that the internal representations of different formats gradually converge as layer depth increases. This suggests that while early layers process format-specific features, deeper layers extract format-agnostic semantic information. This convergence may explain why LLMs can achieve reasonable performance across different table formats. As shown in Table~\ref{tab:comprehensive_results}, the performance gap between Markdown and HTML formats is minimal across all models (typically within 1-2\%), empirically confirming that LLMs can effectively handle both formats despite their different syntactic structures.

\begin{table}[t]
\centering
\caption{Average performance over 2,000 test samples under different input formats and inference strategies.}
\resizebox{\columnwidth}{!}{
\begin{tabular}{lcccc}
\toprule
 & \multicolumn{2}{c}{\textbf{Markdown}} & \multicolumn{2}{c}{\textbf{HTML}} \\
\cmidrule(lr){2-3} \cmidrule(lr){4-5}
\textbf{Model} & Vanilla & CoT & Vanilla & CoT \\
\midrule

\multicolumn{5}{l}{\cellcolor[rgb]{0.957,0.957,0.957}\textit{\textbf{Instruct LLMs}}} \\
\addlinespace[0.3em]
Qwen2.5-7B & 62.15 & 67.20 & 61.35 & 67.80 \\
Llama3.1-8B & 56.95 & 62.35 & 57.95 & 58.35 \\
\midrule

\multicolumn{5}{l}{\cellcolor[rgb]{0.957,0.957,0.957}\textit{\textbf{Table-Specific Fine-tuned LLMs}}} \\
\addlinespace[0.3em]
Table-R1-7B & 69.75 & \textbf{82.30} & 70.30 & \textbf{81.60} \\
Table-R1-8B & 70.85 & 82.25 & 69.90 & 80.60 \\
\midrule

\multicolumn{5}{l}{\cellcolor[rgb]{0.957,0.957,0.957}\textit{\textbf{MoE LLMs}}} \\
\addlinespace[0.3em]
Qwen3-30B-A3B & 66.30 & 70.15 & 68.35 & 72.10 \\
\bottomrule
\end{tabular}}
\label{tab:comprehensive_results}
\end{table}

\subsection{How Does Reasoning Strategy Affect Attention?}
\label{sec:rq4_reasoning}

\begin{tcolorbox}[colback=cyan!5!white, colframe=cyan!45!blue!60, title=\textbf{Takeaway 6}]
Chain-of-Thought (CoT) reasoning leads to higher attention to table content in middle and late layers compared to direct answering, an effect that persists even when analyzing only final answer tokens. Table-specific fine-tuning further amplifies this effect.
\end{tcolorbox}

We compare the attention patterns under direct answering and Chain-of-Thought (CoT) prompting for both vanilla LLMs and their table-tuned variants. As shown in Figure~\ref{reasoning_settings_analysis} (bottom), CoT reasoning leads to a higher proportion of attention allocated to table content, especially in the middle and late layers. Comparing vanilla and table-tuned models (Figure~\ref{reasoning_settings_analysis}, top), table-specific fine-tuning generally increases attention to tables across all layers, with RL-tuned models showing the largest increase when combined with CoT. The performance results in Table~\ref{tab:comprehensive_results} corroborate this: CoT consistently improves accuracy across all models, with particularly striking gains for table-tuned models (e.g., Table-R1-7B improves from 69.75\% to 82.30\%).\footnote{All CoT vs.\ Direct differences are statistically significant (non-overlapping 95\% bootstrap CIs, 1,000 iterations, $N$=2,000). Markdown vs.\ HTML differences fall within the margin of uncertainty for most models. Complete results with confidence intervals are provided in Appendix~\ref{sec:appendix_performance}.}

To further isolate CoT's effect on table grounding from the natural table references within reasoning chains, we recompute the Table Attn Ratio using \textit{only the attention weights at final answer tokens}, excluding all reasoning chain tokens. In early layers---layer 1 for LLaMA-3.1-8B (+1.85\%) and layers 1--4 for Qwen2.5-7B (+13.56\%, +9.12\%, +1.52\%, +7.45\%)---CoT maintains a substantially higher Table Attn Ratio than direct answering even under this restricted setting, confirming that CoT genuinely induces stronger table grounding at the moment of answer generation. In middle-to-late layers, the final-answer-only ratio under CoT falls slightly below that of direct answering, suggesting a division of labor: the reasoning chain has already distilled relevant table information into the residual stream, allowing later layers to attend to the nearby reasoning trajectory rather than re-attending to distant table tokens.

\section{Conclusion}
\label{conclusion}
We present the first systematic empirical study on the internal mechanisms of LLM-based table understanding across 16 LLMs. Our analysis reveals a coherent three-phase workflow: early layers broadly encode the table, middle layers precisely localize query-relevant cells, and late layers amplify the focused content for answer generation. This pattern manifests consistently across architectures—MoE models mirror this through table-specific expert activation in middle layers, while CoT reasoning and table-specific fine-tuning further enhance table engagement. These interconnected findings offer actionable guidance: from optimal input configurations, to inference-time interventions in later layers, to targeted optimization of table-specialized experts.

\section{Limitations}
\label{limitations}
Though this paper presents a systematic empirical study of the internal mechanisms of LLM-based table understanding, there are certain limitations and promising directions that deserve future research. (1) Analyzing LLMs of larger scales and different series. Due to resource limitations, we mainly analyze advanced open-source models with parameters below 32B. Models of large scales such as DeepSeek-V3-671B and Qwen3-235B-A22B also deserves further investigation to analyze the scaling effect on table understanding. (2) Extending to MLLMs. Previous work and recent proprietary models have demonstrated that multimodal LLMs (MLLMs) also possess strong table understanding ability based on table images~\cite{zheng-etal-2024-table-llava,deng-etal-2024-tables_as_images_or_texts,zhou-etal-2025-texts_or_images}. As a result, it is worthwhile to extend our analysis to MLLM scope to discover unique characteristics of multimodal table understanding. (3) Analyzing more tabular tasks. This study primarily focuses on table question answering and table fact verification, two most typical tasks of existing studies. However, it is pertinent to investigate whether our findings can generalize to more complex tabular tasks, such as data analysis and table summarization. (4) Enriching table formats and reasoning strategies. We do not intend to exhaust every possible table formats such as CSV and JSON formats and every reasoning strategies such as program-of-thoughts with in-context examples, which could be further explored by future follow-ups. 

\section{Ethical Considerations}
\label{ethical_considerations}
Our empirical study and all experiments are conducted based 4 public academic benchmarks, which are free and open-source data for research use. The analyzed LLMs are also open-source models that we downloaded from their official websites. As a result, the authors foresee no ethical concerns.


\bibliography{anthology}

\appendix
\label{appendix}

\section{Background}

\subsection{Task Formalization}

Given a table $\mathcal{T}$ and a natural language question $\mathcal{Q}$, Table Question Answering (TableQA) requires the model $G(\cdot)$ to generate a final answer $\mathcal{A}$ based on the above information. Within the context of Large Language Models (LLMs), the input sequence $x$ is composed of three segments serving distinct semantic roles:

\begin{itemize}
    \item \textbf{System Prompt Segment ($\mathcal{S}_{sys}$)}: Comprises the system instructions and model-specific conversation templates (e.g., \texttt{<|begin\_of\_text|>} for Llama-3.1-8B-Instruct ~\citep{llama3_model}).
    \item \textbf{Table Content Segment ($\mathcal{S}_{tab}$)}: Contains the table data serialized in a specific format (e.g., Markdown or HTML).
    \item \textbf{Question Description Segment ($\mathcal{S}_{qst}$)}: Contains the specific user query $\mathcal{Q}$.
\end{itemize}

Let $\mathcal{I}_{sys}$, $\mathcal{I}_{tab}$, and $\mathcal{I}_{q}$ denote the sets of indices corresponding to these segments in the input sequence, the total input index set is $\mathcal{I}_{in} = \mathcal{I}_{sys} \cup \mathcal{I}_{tab} \cup \mathcal{I}_{q}$. The set of indices for tokens generated by the model is denoted as $\mathcal{I}_{gen}$.

\subsection{Architecture of LLMs}
The input sequence is processed by the LLM through $L$ transformer blocks~\citep{vaswani2017attention}. The update process of residual stream at each layer $l$ can be unified as follows:
\begin{align}
    \hat{x}^l &= \text{MHA}^l(\text{LN}(x^{l-1})) + x^{l-1} \\
    x^l &= \text{Trans}^l(\text{LN}(\hat{x}^l)) + \hat{x}^l
\end{align}

where $x^l \in \mathbb{R}^{n \times d}$ represents the hidden state at layer $l$, $d$ is the embedding dimension, and $n$ is the total sequence length. $\text{MHA}$ denotes the Multi-Head Attention module, and $\text{Trans}$ represents the transition layer, which is a Feed-Forward Network (FFN) in dense LLMs or a Mixture-of-Experts (MoE) layer in MoE LLMs.

In the $l$-th layer, the MHA mechanism allocates attention weights by computing the similarity between Queries ($Q$) and Keys ($K$):

\begin{equation}
    \small
    \alpha_{i,j}^{l,h} = \text{softmax} \left( \frac{(x_i^{l-1} W_Q^{l,h})(x_j^{l-1} W_K^{l,h})^T}{\sqrt{d_k}} \right)
\end{equation}
where $\alpha_{i,j}^{l,h}$ denotes the attention weight from token $i$ to token $j$ in the $h$-th attention head of the $l$-th layer. The final output of a single attention head is the aggregation of contributions from all preceding tokens:
\begin{equation}
    \text{MHA}^{l,h}(x_i) = \sum_{j \le i} \alpha_{i,j}^{l,h} (x_j^{l-1} W_{OV}^{l,h})
\end{equation}
Here, $W_{OV}^{l,h} = W_V^{l,h} W_O^{l,h}$ represents the joint value-output projection matrix, which defines the specific path through which information is mapped from input tokens into the residual stream.

For MoE LLMs, the transition layer introduces a sparse activation mechanism where a routing function $G$ selects the Top-$k$ experts for each token. The output is scaled by the routing weights and aggregated:
\begin{equation}
    \text{MoE}(x_i) = \sum_{e \in \text{Top-k}(G_i)} G_{i,e} \cdot E_e(\text{LN}(x_i))
\end{equation}
where $G_i = \text{softmax}(x_i W_G)$ and $E_e$ denotes the $e$-th expert. Through this hierarchical process, the LLM transforms internal representations into the final response.

\section{Interpretability Analysis Methods}
\label{sec:methods}
This section introduces our core methodology for revealing the internal mechanisms of LLMs during tabular reasoning, categorized into three dimensions: \textit{Inner Prompt}, \textit{Inner Table}, and \textit{Contribution}.

\subsection{Inner Prompt: Segment Attention Ratio}

To quantify the allocation of attention resources across different input segments during inference, we define the \textbf{segment attention ratio $\mathcal{D}$}, which measures the degree of attention weight assigned to a specific segment $\mathcal{S}$ at layer $l$, averaged across all attention heads and the entire generation sequence $\mathcal{I}_{gen}$:
\begin{equation}
    \mathcal{D}^l(\mathcal{S}) = \frac{1}{|\mathcal{I}_{gen}| \cdot H} \sum_{i \in \mathcal{I}_{gen}} \sum_{h=1}^H \sum_{j \in \mathcal{S}} \hat{\alpha}_{i,j}^{l,h}
\end{equation}
where $\hat{\alpha}_{i,j}^{l,h} = \frac{\alpha_{i,j}^{l,h}}{\sum_{k \in \mathcal{I}_{in}} \alpha_{i,k}^{l,h}}$ represents the \textbf{rescaled weights for the input segment $\mathcal{I}_{in}$ at generation step $i \in \mathcal{I}_{gen}$}. By observing the trend of $\mathcal{D}^l(\mathcal{S})$ across layers, we can identify the dynamic transition of attention from the structured background ($\mathcal{S}_{tab}$) to the task objective ($\mathcal{S}_{q}$).

\subsection{Inner Table: Table Attention Entropy}

To qualitatively describe the degree of focus on relevant cells within the table, we define the \textbf{average table attention entropy} at layer $l$ as:
\begin{equation}
    \small
    \mathcal{H}^l_{table} = \frac{1}{|\mathcal{I}_{gen}|} \sum_{i \in \mathcal{I}_{gen}} \left( - \sum_{j \in \mathcal{I}_{tab}} \bar{P}^l_i(j) \log \bar{P}^l_i(j) \right)
\end{equation}
where the inner-table attention distribution $\bar{P}^l_i(j)$ is defined as:
\begin{equation}
    \small
    \bar{P}^l_i(j) = \frac{\sum_{h=1}^H \alpha_{i,j}^{l,h}}{\sum_{k \in \mathcal{I}_{tbl}} \sum_{h=1}^H \alpha_{i,k}^{l,h}}, \quad \forall j \in \mathcal{I}_{tab}
\end{equation}

A higher entropy indicates a uniform distribution of attention across the table, whereas a lower entropy signifies that attention is concentrated in a few specific cells. This implies that the layer may identify the task-relevant information.

\subsection{Value-Weighted Contribution}

Following the perspective of previous researches~\citep{kobayashi-etal-2020-attention,gu2024attention,kang2025see} that attention weights alone do not quantify the actual magnitude of change in the residual stream, we introduce the \textbf{Value-Weighted Contribution} analysis. To capture the model's behavior throughout the generation process, we compute the averaged contribution over $\mathcal{I}_{gen}$:

\begin{equation}
    \small
    C^l(\mathcal{S}) = \frac{1}{|\mathcal{I}_{gen}| \cdot H} \sum_{i \in \mathcal{I}_{gen}} \left\| \sum_{j \in \mathcal{S}} \sum_{h=1}^H \alpha_{i,j}^{l,h} x_{j}^{l-1} W_{OV}^{l,h} \right\|_2
\end{equation}

where $\mathcal{S} \in \{\mathcal{I}_{sys}, \mathcal{I}_{tab}, \mathcal{I}_{q}\}$. By integrating the hidden states $x_j$ with the projection matrix $W_{OV}$, this formula faithfully reconstructs the intensity of the information volume injected by a specific segment $\mathcal{S}$ at layer $l$ to drive the following reasoning.

\section{Experimental Details}
\label{sec:experimental_details}

\subsection{Prompt Templates}
\label{prompt_template}

We provide the complete prompt templates used in our experiments. The input structure follows the standard practice of prompting LLMs for table tasks, consisting of three segments: (1) a prompt template, (2) a table content segment, and (3) a user question segment.

\begin{center}
\begin{tcolorbox}[
    colback=blue!5!white,
    colframe=blue!55!black,
    width=0.48\textwidth,
    title={\textbf{Overall Prompt Template}}
]
{
\textbf{System Prompt:} \\
\texttt{You are a helpful assistant.} \\

\textbf{User Prompt:} \\
\texttt{TABLE:} \\
\texttt{\{table\_content\}} \\
\texttt{INPUT:} \\
\texttt{\{question\_content\}}
}
\end{tcolorbox}
\end{center}

\begin{tcolorbox}[colback=black!3!white, colframe=black!70!white, title={Prompt Template for Table Question Answering (WTQ, HiTab, AITQA)}, fontupper=\footnotesize, fonttitle=\footnotesize]
\textbf{TABLE:} \\
\texttt{\{table\_content (markdown / html)\}} \\

\textbf{INPUT:} \\
\texttt{Given the table titled `\{table\_title\}', answer the following question based on the given table. The final answer should be concise and use the following format:}
\begin{verbatim}
```json
{
    "answer": ["answer1", "answer2", ...]
}
```
\end{verbatim}
\texttt{Question: \{question\_content\}}
\end{tcolorbox}

\begin{tcolorbox}[colback=black!3!white, colframe=black!70!white, title=Prompt Template for Table Fact Verification (TabFact),
fontupper=\footnotesize, fonttitle=\footnotesize]
\textbf{TABLE:} \\
\texttt{\{table\_content (markdown / html)\}} \\

\textbf{INPUT:} \\
\texttt{Given the table titled `\{table\_title\}', determine whether the following statement is entailed or refuted by the given table (Output 1 for entailed and 0 for refuted):} \\
\texttt{Statement: \{statement\_content\}}
\end{tcolorbox}

\subsection{Table Serialization Formats}

Unless otherwise specified, we use Markdown as the default table serialization format. Figure~\ref{fig:serialization_example} illustrates the two serialization formats (Markdown and HTML) used in our experiments.

\begin{figure}[h]
\centering
\begin{minipage}{0.48\textwidth}
\begin{tcolorbox}[colback=green!5!white, colframe=green!50!black, title=Markdown Format, fontupper=\scriptsize, fonttitle=\footnotesize]
\begin{verbatim}
| Name  | Age | City    |
|-------|-----|---------|
| Alice | 25  | Beijing |
| Bob   | 30  | Shanghai|
\end{verbatim}
\end{tcolorbox}
\end{minipage}
\hfill
\begin{minipage}{0.48\textwidth}
\begin{tcolorbox}[colback=orange!5!white, colframe=orange!50!black, title=HTML Format, fontupper=\scriptsize, fonttitle=\footnotesize]
\begin{verbatim}
<table>
  <tr><th>Name</th><th>Age</th>
      <th>City</th></tr>
  <tr><td>Alice</td><td>25</td>
      <td>Beijing</td></tr>
  <tr><td>Bob</td><td>30</td>
      <td>Shanghai</td></tr>
</table>
\end{verbatim}
\end{tcolorbox}
\end{minipage}
\caption{Examples of table serialization formats.}
\label{fig:serialization_example}
\end{figure}

\subsection{Analyzed Models}
\label{appendix::used_llms}

To provide a comprehensive analysis across different model architectures and training paradigms, we categorize the investigated LLMs into three groups:

\paragraph{(1) Instruct LLMs.}
We analyze general-purpose instruction-tuned models that represent the mainstream LLM paradigm, including the Llama series~\citep{llama3_model} (e.g., Llama-3.1-8B-Instruct) and the Qwen series~\citep{qwen2_5} (e.g., Qwen-2.5-7B-Instruct).

\paragraph{(2) Table-Specific Fine-tuned LLMs.}
To understand how specialized training affects internal mechanisms, we include models specifically optimized for tabular tasks. We further distinguish between \textit{SFT-only} models trained through supervised fine-tuning (e.g., the TAMA series~\citep{tama}) and \textit{RL-only} models optimized via reinforcement learning (e.g., the Table-R1-Zero series~\citep{table_r1_zero}).

\paragraph{(3) Mixture-of-Experts (MoE) LLMs.}
Beyond dense architectures, we extend our analysis to MoE models to examine whether sparse expert activation leads to different table understanding behaviors. Representative models include OLMoE-1B-7B~\citep{olmoe}, DeepSeek-V2-Lite-Chat~\citep{deepseek_v2}, and Qwen3-30B-A3B~\citep{qwen3}.

\section{Model Analysis Details}
\label{sec:appendix}

\subsection{Benchmark Performance}
\label{sec:appendix_performance}


\begin{table*}[htbp]
\centering
\begin{tabular}{lcccc}
\toprule
\textbf{Model} & \textbf{WTQ} & \textbf{TabFact} & \textbf{HiTab} & \textbf{AITQA} \\
\midrule

\multicolumn{5}{l}{\cellcolor[rgb]{0.957,0.957,0.957}\textit{\textbf{Instruct LLMs}}} \\
\addlinespace[0.3em]
LLaMA3.2-1B-Instruct & 6.00 & 5.40 & 6.40 & 9.80 \\
LLaMA3.2-3B-Instruct & 33.00 & 58.00 & 30.80 & 52.00 \\
LLaMA3.1-8B-Instruct & 50.20 & 70.80 & 49.00 & 57.80 \\
Qwen2.5-3B-Instruct & 52.40 & 78.00 & 49.80 & 55.60 \\
Qwen3-4B-Instruct-2507 & 52.00 & 75.20 & 49.60 & 60.00 \\
Qwen2.5-7B-Instruct & 51.00 & 73.00 & 53.40 & 71.20 \\
Qwen2.5-14B-Instruct & 56.40 & \textbf{88.60} & 55.80 & 74.40 \\
Qwen2.5-32B-Instruct & \textbf{64.60} & 85.80 & 68.80 & 83.60 \\
\midrule 

\multicolumn{5}{l}{\cellcolor[rgb]{0.957,0.957,0.957}\textit{\textbf{Table-Specific Fine-tuned LLMs}}} \\
\addlinespace[0.3em]
TAMA-Qwen2.5 & 52.40 & 74.80 & 56.80 & 71.60 \\
TAMA & 50.60 & 72.60 & 66.80 & 82.00 \\
TableGPT2 & 56.80 & 74.80 & 59.80 & 84.20 \\
Table-R1-7B-Zero & 53.80 & 76.20 & 63.80 & \textbf{85.20} \\
Table-R1-8B-Zero & 60.40 & 70.40 & \textbf{74.00} & 78.60 \\
\midrule 

\multicolumn{5}{l}{\cellcolor[rgb]{0.957,0.957,0.957}\textit{\textbf{MoE LLMs}}} \\
\addlinespace[0.3em]
OLMoE-1B-7B-Instruct & 6.82 & 2.20 & 5.01 & 14.60 \\
DeepSeek-V2-Lite-Chat & 28.60 & 63.40 & 30.40 & 53.00 \\
Qwen3-30B-A3B-Instruct-2507 & 56.40 & 78.20 & 57.40 & 73.20 \\
\bottomrule
\end{tabular}
\caption{LLMs' performance (accuracy \%) on evaluation data of four benchmarks. Best results are bolded.}
\label{tab:main_results}
\end{table*}

Table~\ref{tab:main_results} reports the accuracy of all analyzed LLMs on our four evaluation benchmarks. Several observations emerge: (1) larger models generally achieve better performance; (2) table-specific fine-tuned models (TAMA, TableGPT2, Table-R1) show notable improvements on hierarchical table datasets (HiTab, AITQA); and (3) RL-tuned models (Table-R1 series) demonstrate particularly strong performance on complex table structures.

Table~\ref{tab:comprehensive_results_ci} reports the performance of representative models under different input format and inference strategy combinations, with 95\% bootstrap confidence intervals (1,000 iterations, $N$=2,000 per condition). All CoT vs.\ Direct differences are statistically significant (non-overlapping CIs), confirming the reliability of prompting strategy conclusions. Markdown vs.\ HTML differences fall within the margin of uncertainty for most models, consistent with the observation that format-specific differences in early layers gradually converge in deeper layers, resulting in minimal overall performance gaps.

\begin{table}[t]
\centering
\caption{Performance (\%) with 95\% bootstrap confidence intervals under different input formats and inference strategies. $\dagger$ marks statistically significant differences between CoT and Direct (non-overlapping CIs).}
\resizebox{\columnwidth}{!}{
\begin{tabular}{lcccc}
\toprule
 & \multicolumn{2}{c}{\textbf{Markdown}} & \multicolumn{2}{c}{\textbf{HTML}} \\
\cmidrule(lr){2-3} \cmidrule(lr){4-5}
\textbf{Model} & Direct & CoT & Direct & CoT \\
\midrule
\multicolumn{5}{l}{\cellcolor[rgb]{0.957,0.957,0.957}\textit{\textbf{Instruct LLMs}}} \\
\addlinespace[0.3em]
Qwen2.5-7B   & 62.2\,{\small$\pm$2.1} & 67.2\,{\small$\pm$2.0}$^\dagger$ & 61.4\,{\small$\pm$2.1} & 67.8\,{\small$\pm$2.0}$^\dagger$ \\
Llama-3.1-8B & 57.0\,{\small$\pm$2.2} & 62.4\,{\small$\pm$2.1}$^\dagger$ & 58.0\,{\small$\pm$2.1} & 58.4\,{\small$\pm$2.1}$^\dagger$ \\
\midrule
\multicolumn{5}{l}{\cellcolor[rgb]{0.957,0.957,0.957}\textit{\textbf{Table-Specific Fine-tuned LLMs}}} \\
\addlinespace[0.3em]
Table-R1-7B  & 69.8\,{\small$\pm$2.0} & 82.3\,{\small$\pm$1.7}$^\dagger$ & 70.3\,{\small$\pm$1.9} & 81.6\,{\small$\pm$1.7}$^\dagger$ \\
Table-R1-8B  & 70.8\,{\small$\pm$2.0} & 82.2\,{\small$\pm$1.7}$^\dagger$ & 69.9\,{\small$\pm$2.1} & 80.6\,{\small$\pm$1.8}$^\dagger$ \\
\midrule
\multicolumn{5}{l}{\cellcolor[rgb]{0.957,0.957,0.957}\textit{\textbf{MoE LLMs}}} \\
\addlinespace[0.3em]
Qwen3-30B-A3B & 66.3\,{\small$\pm$2.1} & 70.2\,{\small$\pm$2.0}$^\dagger$ & 68.3\,{\small$\pm$2.0} & 72.1\,{\small$\pm$1.9}$^\dagger$ \\
\bottomrule
\end{tabular}}
\label{tab:comprehensive_results_ci}
\end{table}

\subsection{Analysis of Effective Depth }
\label{sec:appendix_effective_depth}

\begin{table*}[t]
    \centering
    \begin{tabular}{l c c c}
        \toprule
        \textbf{Model} & \textbf{Total Layers} & \textbf{Eff. Layers} & \textbf{Eff. Ratio} \\
        \midrule
        \multicolumn{4}{l}{\cellcolor[rgb]{0.957,0.957,0.957}\textit{\textbf{Instruct LLMs}}} \\
        \addlinespace[0.3em]
        Llama-3.2-1B-Instruct       & 16 & 14 & 0.88 \\
        Llama-3.2-3B-Instruct       & 28 & 23 & 0.82 \\
        Meta-Llama-3.1-8B-Instruct  & 32 & 25 & 0.78 \\
        Qwen2.5-3B-Instruct         & 36 & 33 & 0.92 \\
        Qwen2.5-7B-Instruct         & 28 & 25 & 0.89 \\
        Qwen2.5-14B-Instruct        & 48 & 44 & 0.92 \\
        Qwen2.5-32B-Instruct        & 64 & 57 & 0.89 \\
        Qwen3-4B-Instruct-2507      & 36 & 24 & 0.67 \\
        \midrule

        \multicolumn{4}{l}{\cellcolor[rgb]{0.957,0.957,0.957}\textit{\textbf{Table-Specific Fine-tuned LLMs}}} \\
        \addlinespace[0.3em]
        TableGPT2-7B                & 28 & 25 & 0.89 \\
        TAMA-QWen2.5                & 28 & 25 & 0.89 \\
        Table-R1-Zero-7B            & 28 & 25 & 0.89 \\
        TAMA-1e-6                   & 32 & 25 & 0.78 \\
        Table-R1-Zero-8B            & 32 & 24 & 0.75 \\
        \midrule

        \multicolumn{4}{l}{\cellcolor[rgb]{0.957,0.957,0.957}\textit{\textbf{MoE LLMs}}} \\
        \addlinespace[0.3em]
        OLMoE-1B-7B-0924-Instruct   & 16 & 14 & 0.88 \\
        DeepSeek-V2-Lite-Chat       & 27 & 21 & 0.78 \\
        Qwen3-30B-A3B-Instruct-2507 & 48 & 45 & 0.94 \\
        \bottomrule
    \end{tabular}
    \caption{Layer effectiveness analysis on table-related tasks using LogitLens. Eff. Layers identifies the first layer where the predicted vocabulary distribution aligns with the final output (defined by the top-5 token overlap first exceeds 0.3). Eff. Ratio (Eff. Layers / Total Layers) reflects the stage at which the model stabilizes its prediction; a lower ratio indicates that the model reaches its final decision in its shallower layers.}
    \label{tab:model_layer_effectiveness}
\end{table*}

\begin{figure*}[t]
  \centering
  \includegraphics[width=\linewidth]{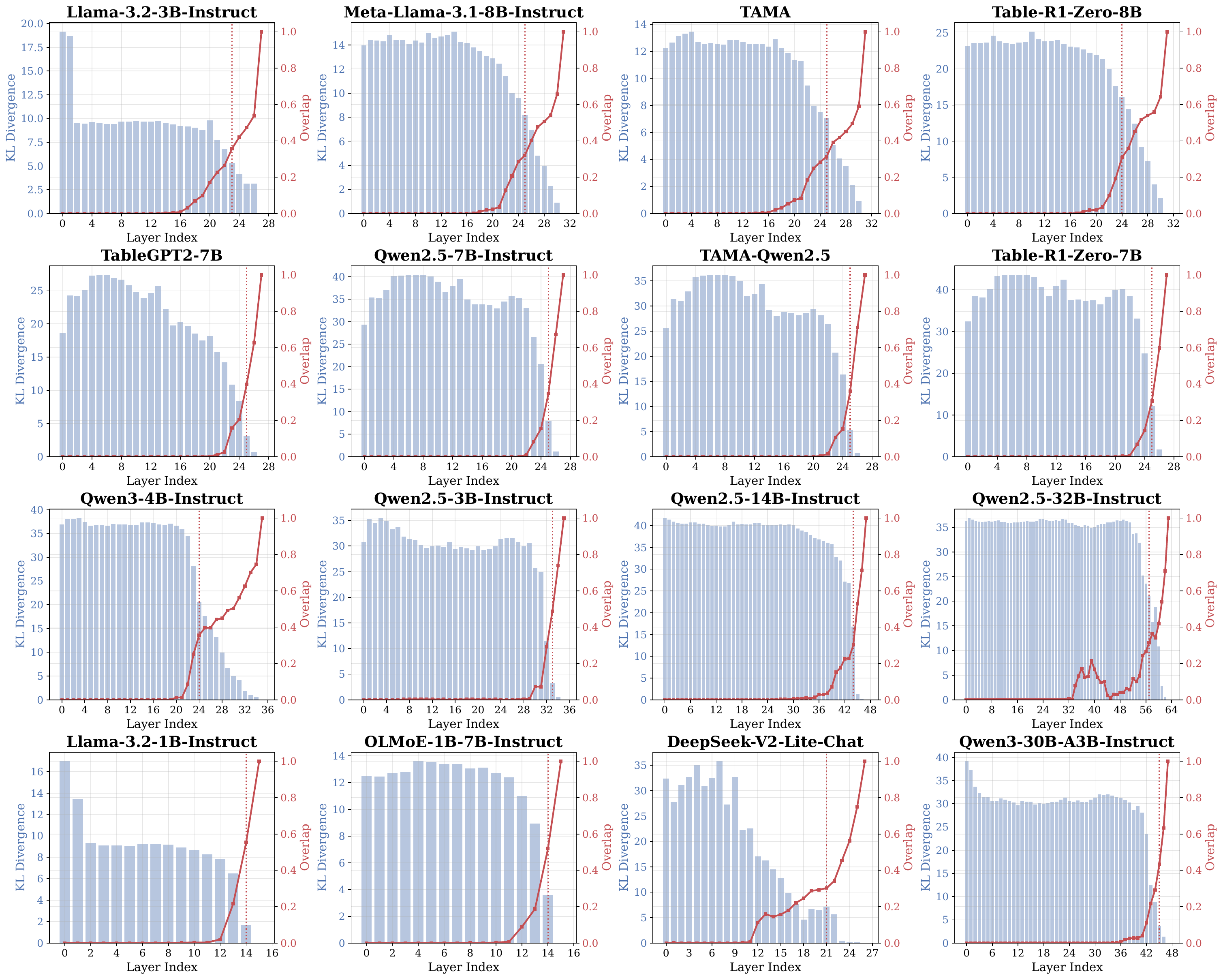}
  \caption{Analysis results of LLMs' effectiveness layer.}
  \label{effective_depth_analysis_final_all}
\end{figure*}

We extend the effective depth analysis from Section~\ref{sec:rq2} to all analyzed models. Table~\ref{tab:model_layer_effectiveness} summarizes the layer effectiveness statistics, while Figure~\ref{effective_depth_analysis_final_all} provides detailed layer-wise prediction stability curves for each model.

As described in the main text, we define the \textit{effective layer} as the first layer where the top-5 token overlap between the intermediate prediction and the final output exceeds 0.3. The \textit{effective ratio} (Eff. Ratio = Eff. Layers / Total Layers) indicates how deeply the model processes before reaching a stable prediction; a lower ratio suggests earlier stabilization.

Key observations from the extended analysis:
\begin{itemize}
    \item Most models stabilize their predictions in the later 75--90\% of their total layers, confirming that table tasks require deep processing.
    \item Table-specific fine-tuning (TAMA, Table-R1) does not significantly alter the effective depth compared to their base models, suggesting that layer-wise functionality is largely determined during pre-training.
    \item MoE models (OLMoE, DeepSeek-V2-Lite, Qwen3-30B-A3B) exhibit similar effective ratios to dense models of comparable capacity.
\end{itemize}

\begin{figure*}[h]
  \centering
  \includegraphics[width=0.8\linewidth]{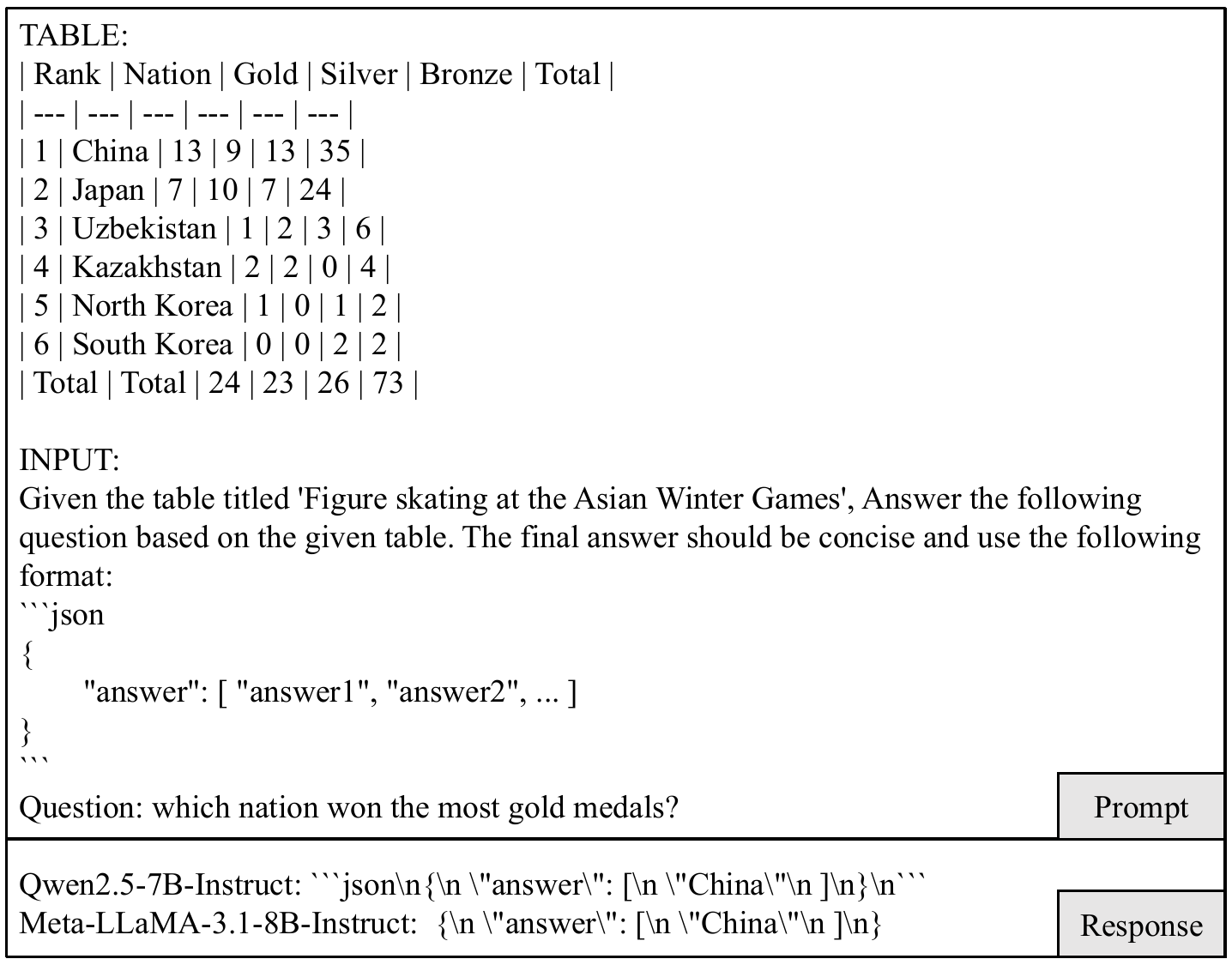}
  \caption{The qualitative case example for Q1.}
\end{figure*}

\section{More Discussions}


\subsection{Connection to Retrieval Heads}
\label{sec:rq1_retrieval}

\begin{figure*}[t]
  \centering
  \includegraphics[width=0.95\linewidth]{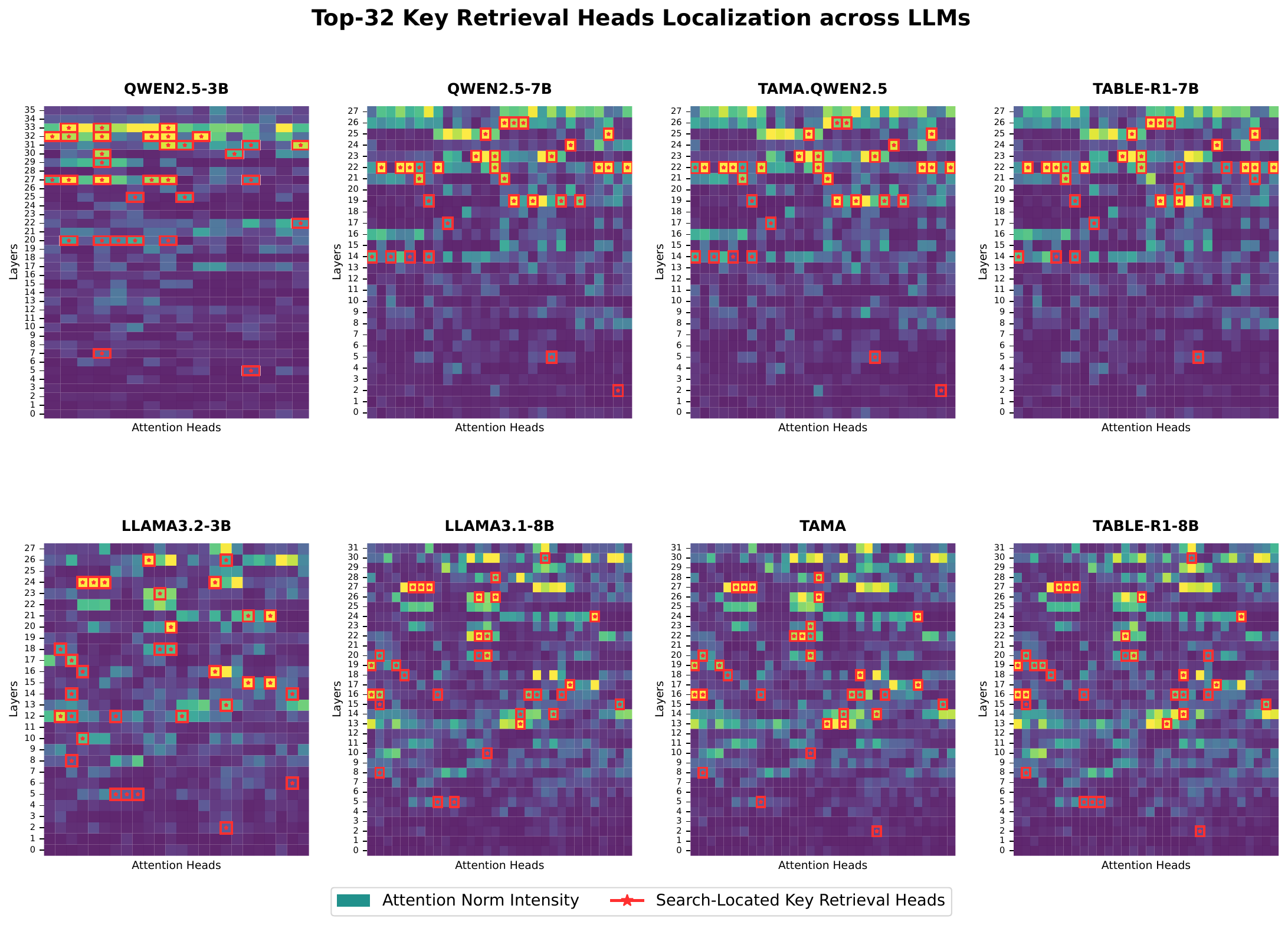}
  \caption{Top-32 retrieval heads' overlap with attention norm heatmaps across LLMs. Background color indicates attention norm intensity (yellow = higher); red boxes mark the Top-32 retrieval heads identified by prior work. The substantial overlap, especially in middle-to-late layers, suggests that table understanding reuses pre-existing retrieval circuits.}
  \label{fig:retrieval_overlap}
\end{figure*}

Recent work has identified \textit{retrieval heads} which is a sparse set of attention heads responsible for copying relevant information from context to output~\citep{wu2024retrieval_head, xiaoduoattention}. Given that table QA inherently requires locating and extracting specific cell values, we hypothesize that table understanding may share similar mechanisms.

To investigate this, we compare the attention heads most critical for table tasks (measured by attention norm intensity) with the Top-32 retrieval heads identified by~\citet{wu2024retrieval_head} using our analysis tabular data. As shown in Figure~\ref{fig:retrieval_overlap}, the retrieval heads (marked with red boxes) predominantly align with regions of high attention norm intensity (yellow areas) across all examined models. This overlap is particularly pronounced in the middle-to-late layers: for instance, in Qwen2.5-7B, retrieval heads cluster around layers 19--27; in LLaMA3.1-8B, they concentrate in layers 14--15 and 26--28; in TAMA and Table-R1-8B, similar patterns emerge in layers 13--17 and 21--28.

Notably, this alignment pattern persists across both vanilla instruction-tuned models (Qwen2.5, LLaMA3) and table-specific fine-tuned models (TAMA, Table-R1), suggesting that table-specific training does not fundamentally alter which heads are responsible for information retrieval—it may instead enhance the efficiency of these pre-existing circuits.

This finding suggests that LLMs leverage their pre-existing retrieval mechanisms for table understanding, rather than developing entirely separate circuits. It also has practical implications: KV cache compression techniques designed for retrieval heads~\citep{xiaoduoattention} may be directly applicable to accelerate table reasoning tasks without significant performance degradation.

\section{Attention Dynamics for All Models}
\label{sec:appendix_attention_all}

To validate the generalizability of our findings in Section~\ref{sec:rq1}, we present the complete attention dynamics analysis for all 16 LLMs examined in this study. Each figure follows the same layout as Figure~\ref{main_analysis_pic}: upper panels show case-level attention patterns at selected layers, while lower panels display aggregated metrics including segment attention ratio, table attention entropy, and attention contribution across all layers. Despite variations in model architecture (dense vs. MoE), scale (1B to 32B), and training paradigm (instruction-tuning, SFT, RL), all models exhibit the consistent three-phase pattern: dispersed attention in early layers, concentrated focus on query-relevant cells in middle layers, and amplified contributions in late layers. Notably, table-specific fine-tuned models (TAMA, Table-R1, TableGPT2) show slightly higher attention ratios to table content compared to their base models, while MoE models (DeepSeek-V2-Lite, OLMoE, Qwen3-30B-A3B) display more gradual entropy transitions across layers.
\begin{figure*}[t]
  \centering
  \includegraphics[width=\linewidth]{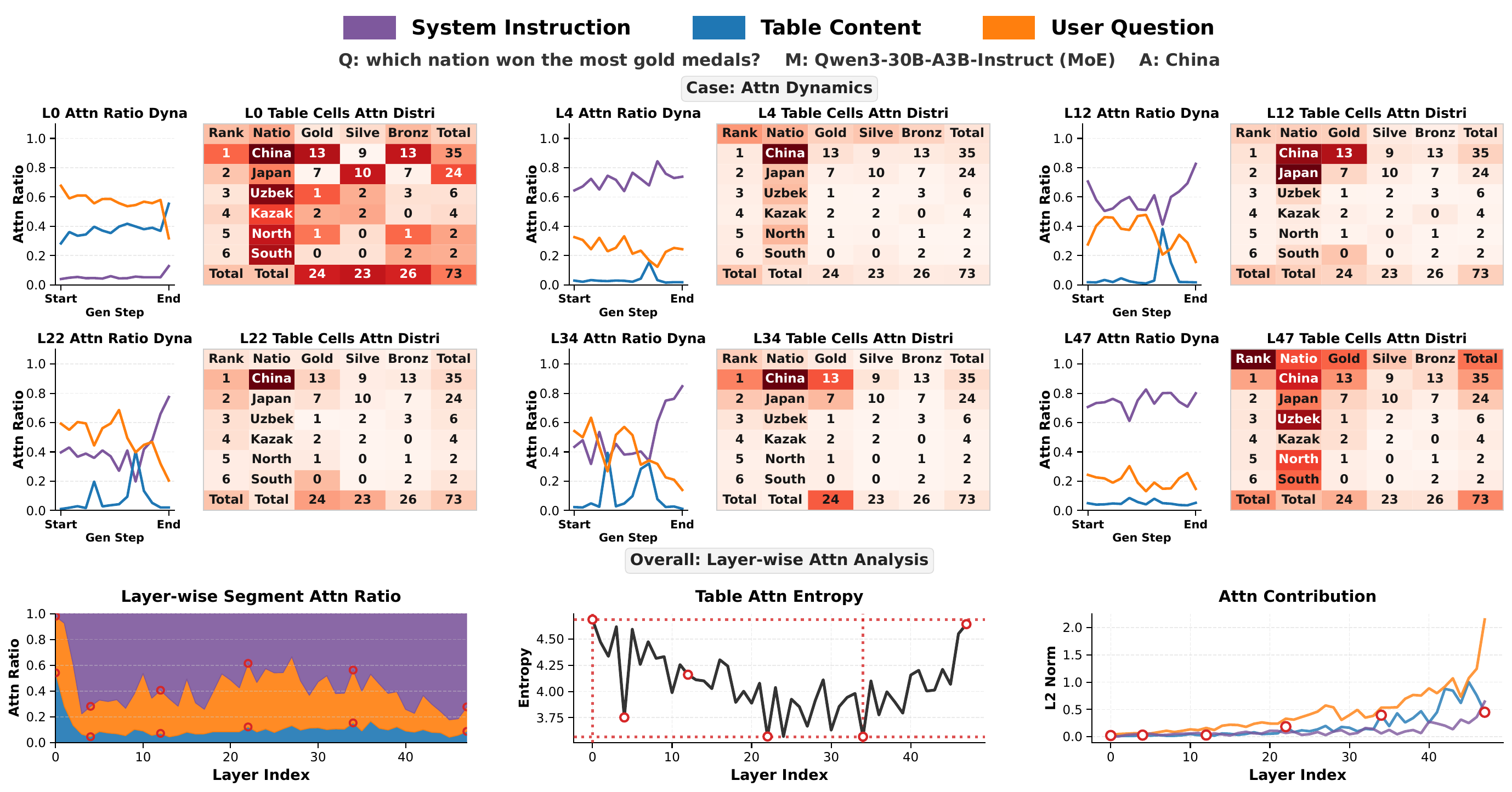}
  \caption{Attention dynamics of Qwen3-30B-A3B-Instruct-2507.}
  \label{llama3_1_main_analysis}
\end{figure*}

\begin{figure*}[t]
  \centering
  \includegraphics[width=\linewidth]{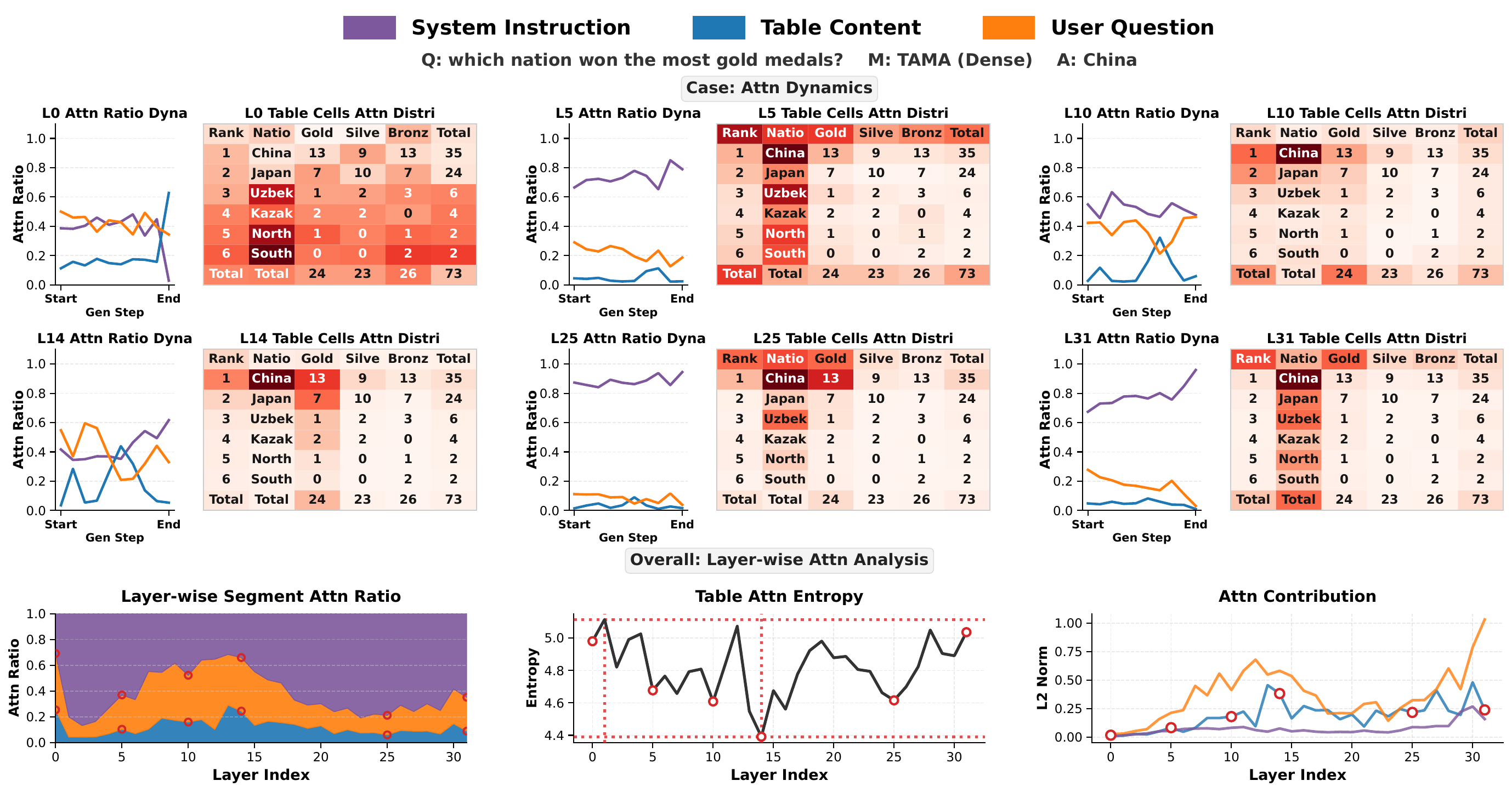}
  \caption{Attention dynamics of TAMA.}
  \label{tama_main_analysis}
\end{figure*}

\begin{figure*}[t]
  \centering
  \includegraphics[width=\linewidth]{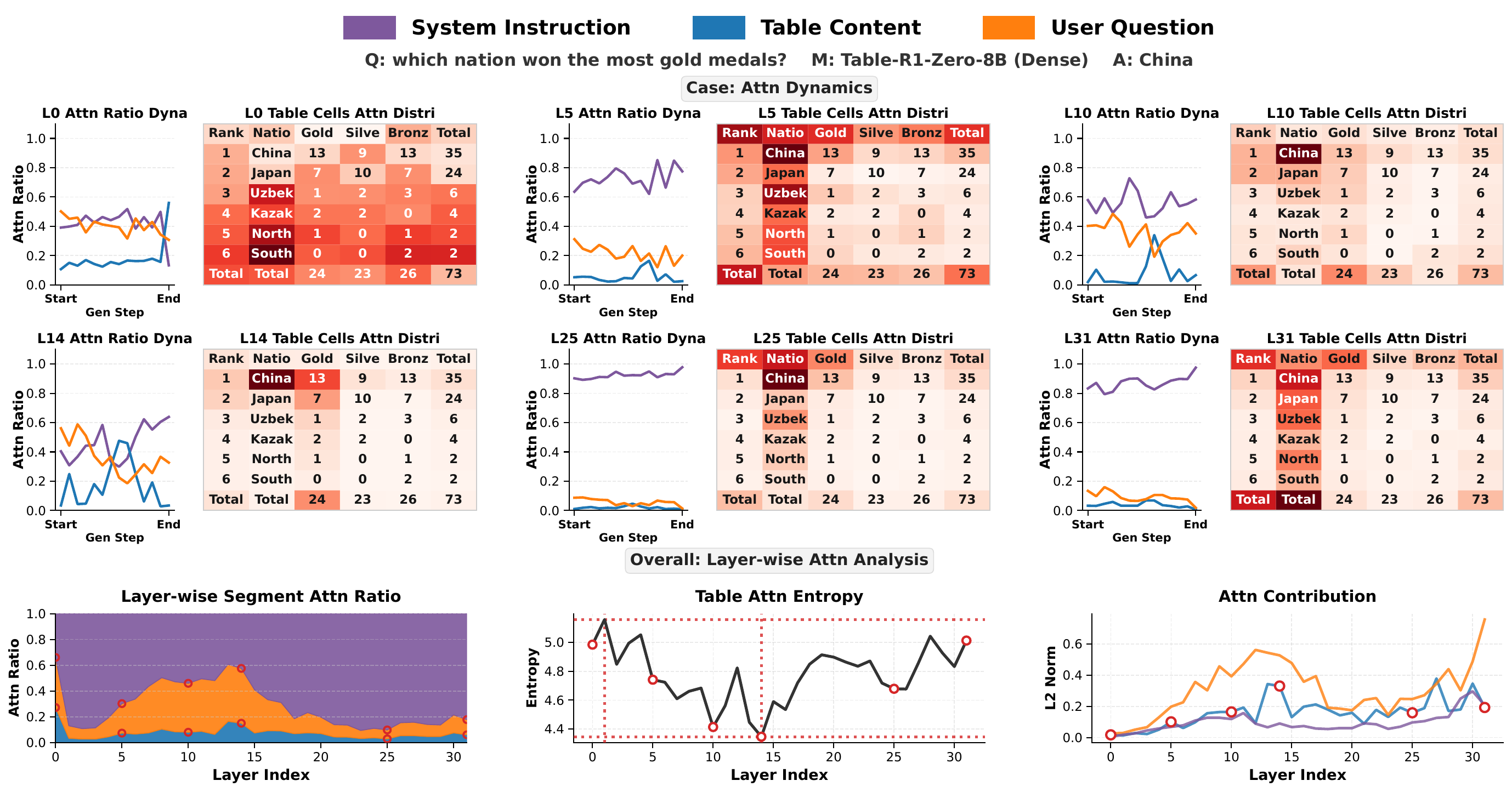}
  \caption{Attention dynamics of Table-R1-Zero-8B.}
  \label{table_r1_8b_main_analysis}
\end{figure*}

\begin{figure*}[t]
  \centering
  \includegraphics[width=\linewidth]{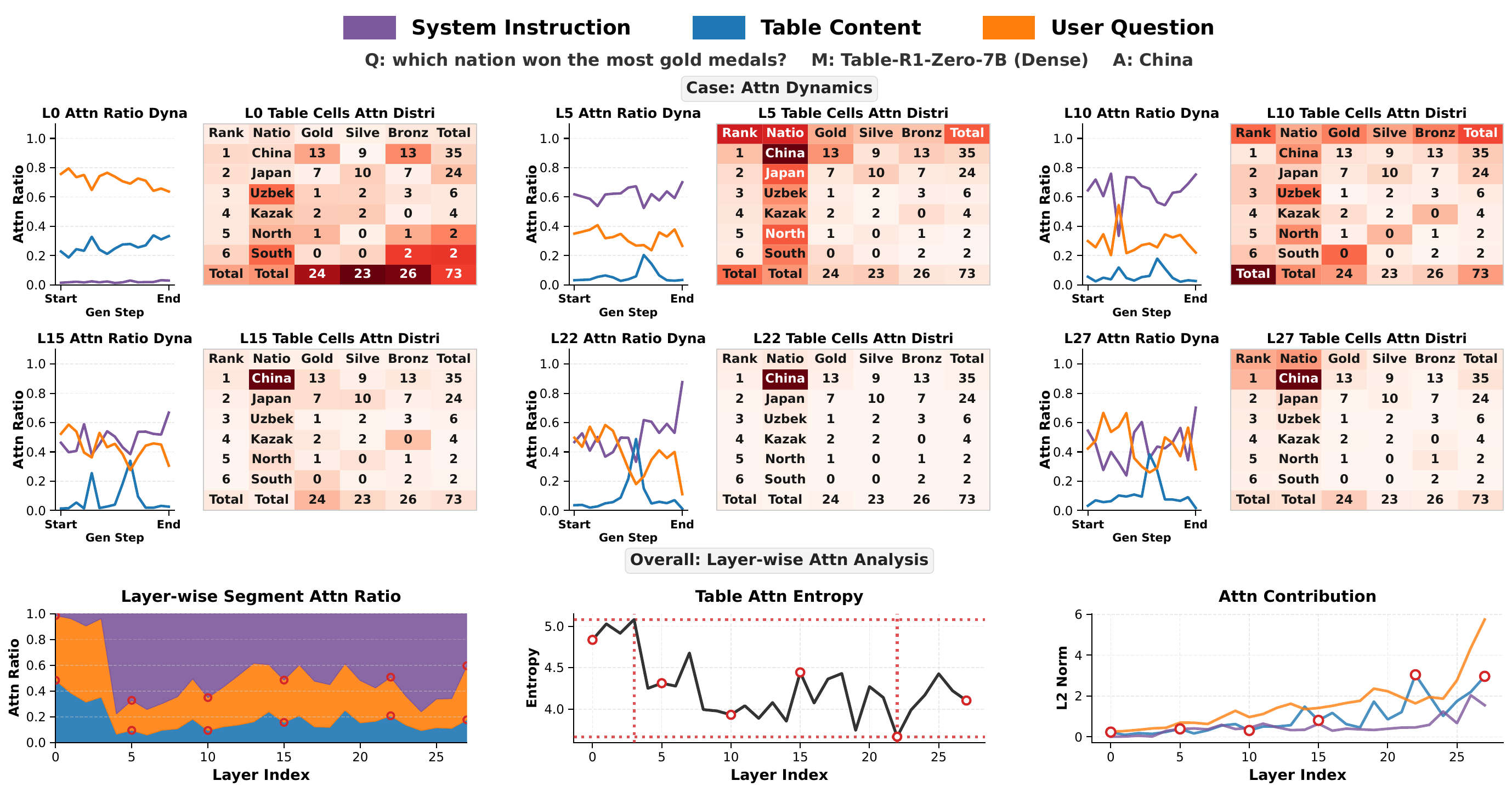}
  \caption{Attention dynamics of Table-R1-Zero-7B.}
  \label{table_r1_7b_main_analysis}
\end{figure*}


\begin{figure*}[t]
  \centering
  \includegraphics[width=\linewidth]{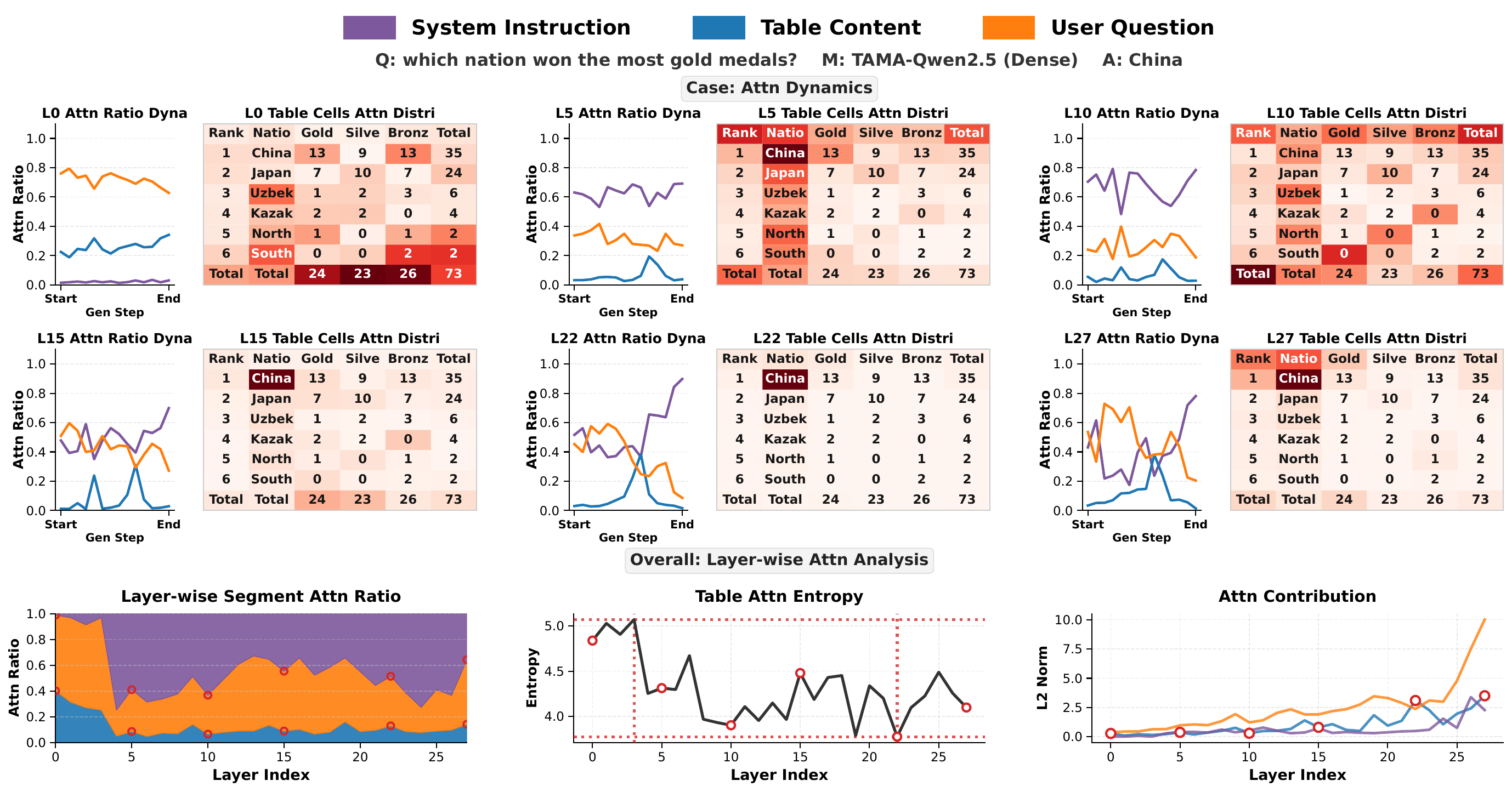}
  \caption{Attention dynamics of TAMA-Qwen2.5.}
  \label{tama_qwen_2_5_main_analysis}
\end{figure*}

\begin{figure*}[t]
  \centering
  \includegraphics[width=\linewidth]{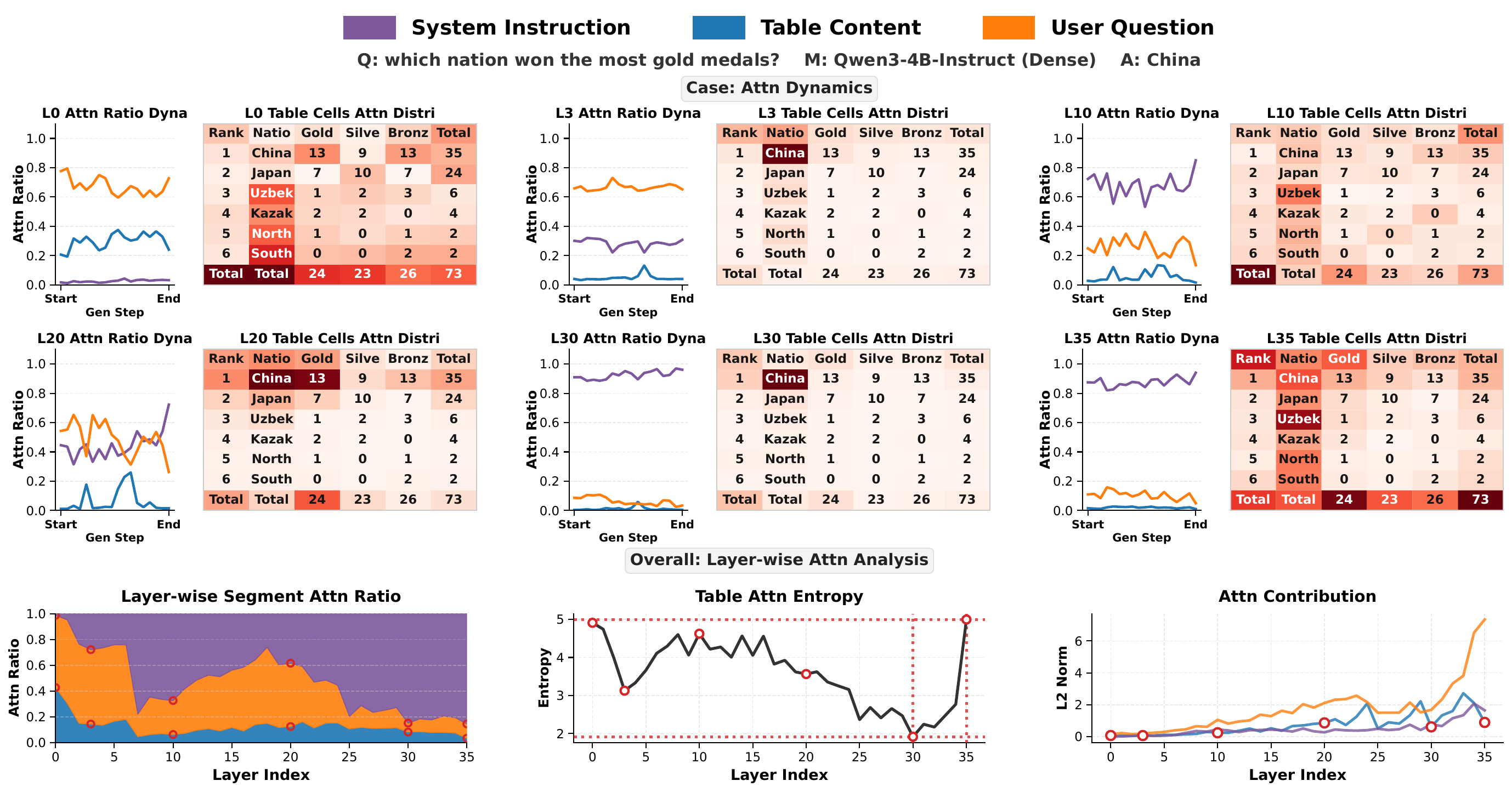}
  \caption{Attention dynamics of Qwen3-4B-Instruct.}
  \label{qwen_3_main_analysis}
\end{figure*}

\begin{figure*}[t]
  \centering
  \includegraphics[width=\linewidth]{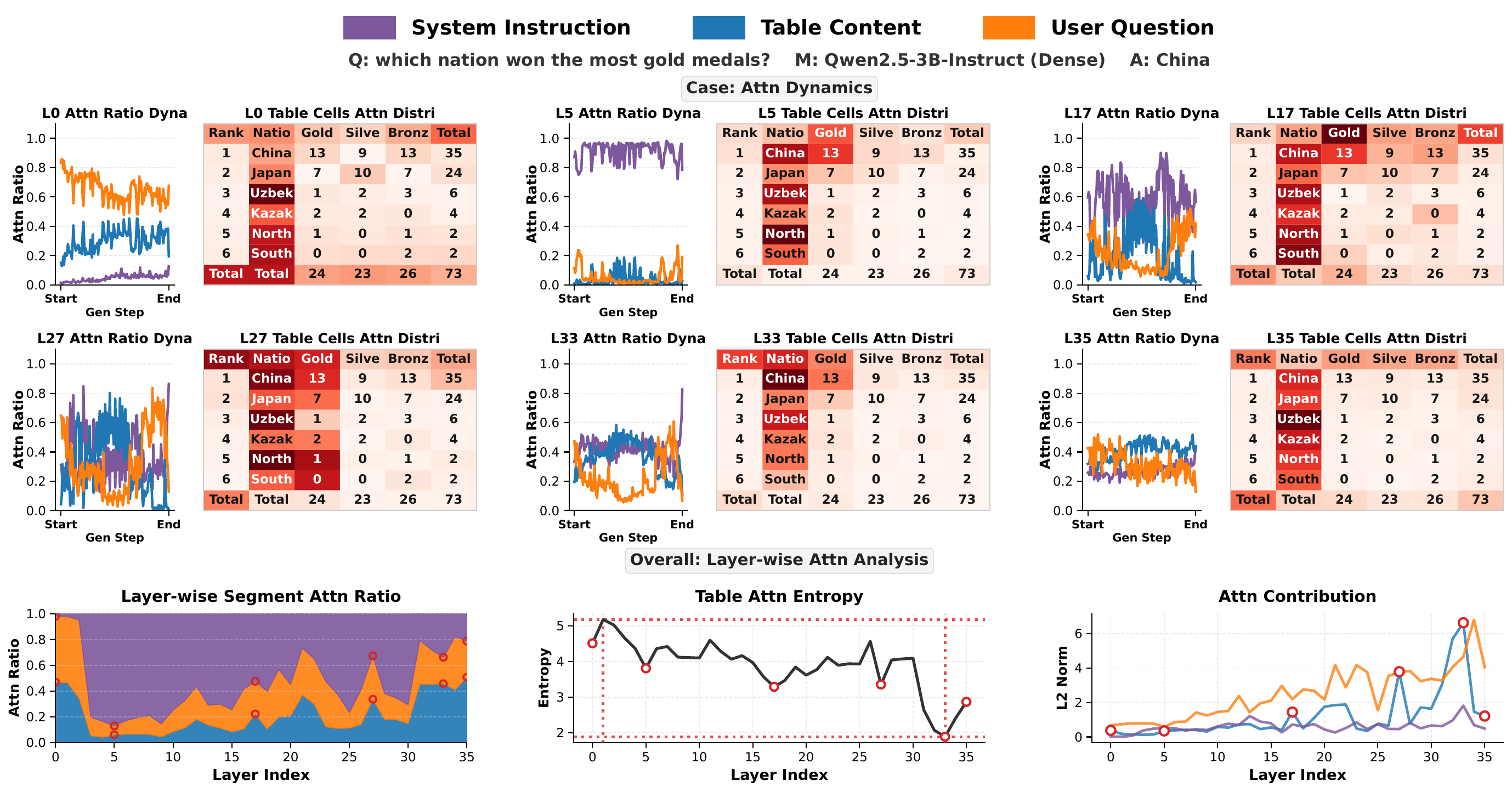}
  \caption{Attention dynamics of Qwen2.5-3B-Instruct.}
  \label{qwen2_5_3b_main_analysis}
\end{figure*}

\begin{figure*}[t]
  \centering
  \includegraphics[width=\linewidth]{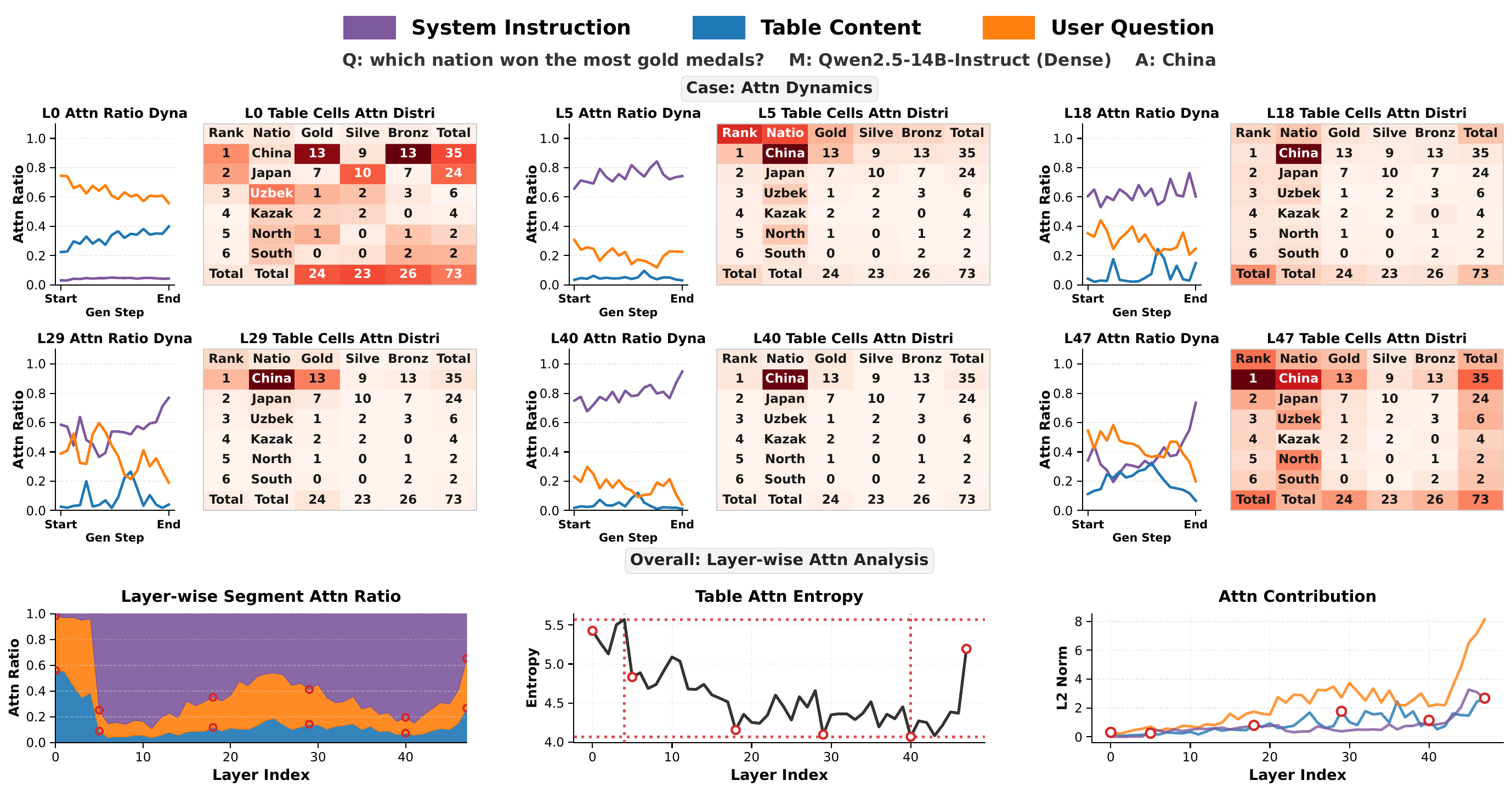}
  \caption{Attention dynamics of Qwen2.5-14B-Instruct.}
  \label{qwen2_5_14b_main_analysis}
\end{figure*}

\begin{figure*}[t]
  \centering
  \includegraphics[width=\linewidth]{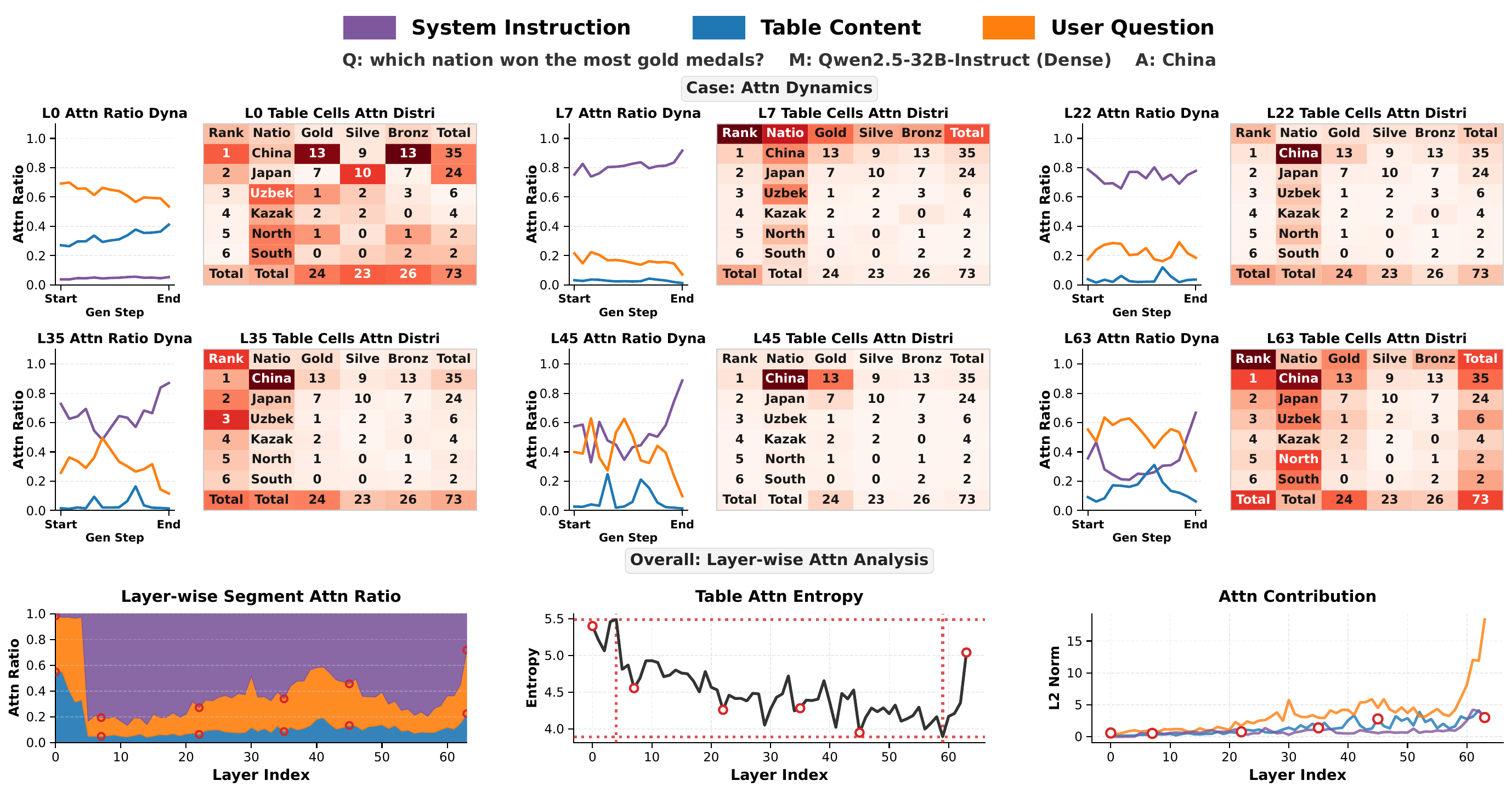}
  \caption{Attention dynamics of Qwen2.5-32B-Instruct.}
  \label{qwen2_5_32b_main_analysis}
\end{figure*}

\begin{figure*}[t]
  \centering
  \includegraphics[width=\linewidth]{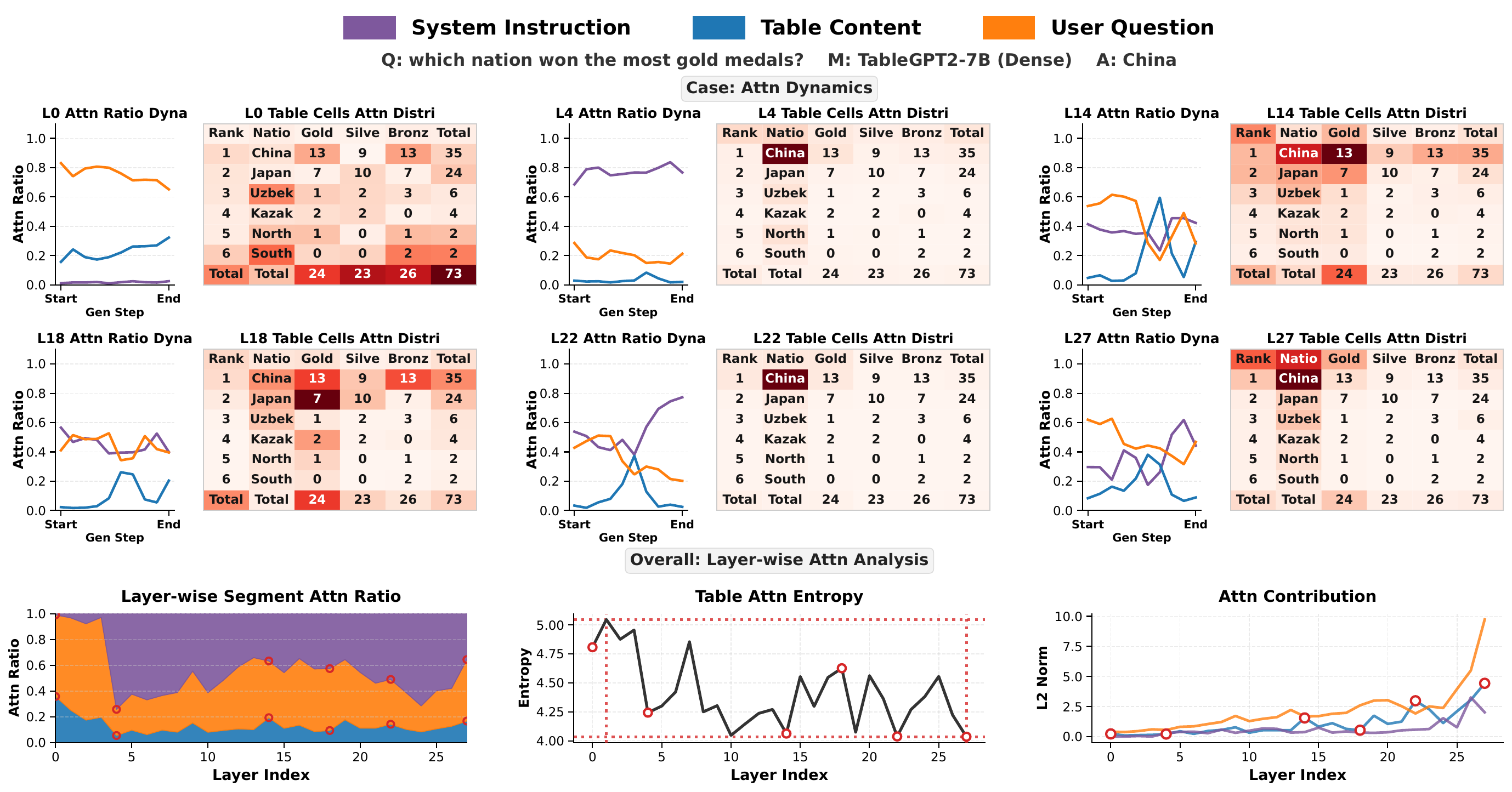}
  \caption{Attention dynamics of TableGPT2-7B.}
  \label{tablegpt2_7b_main_analysis}
\end{figure*}

\begin{figure*}[t]
  \centering
  \includegraphics[width=\linewidth]{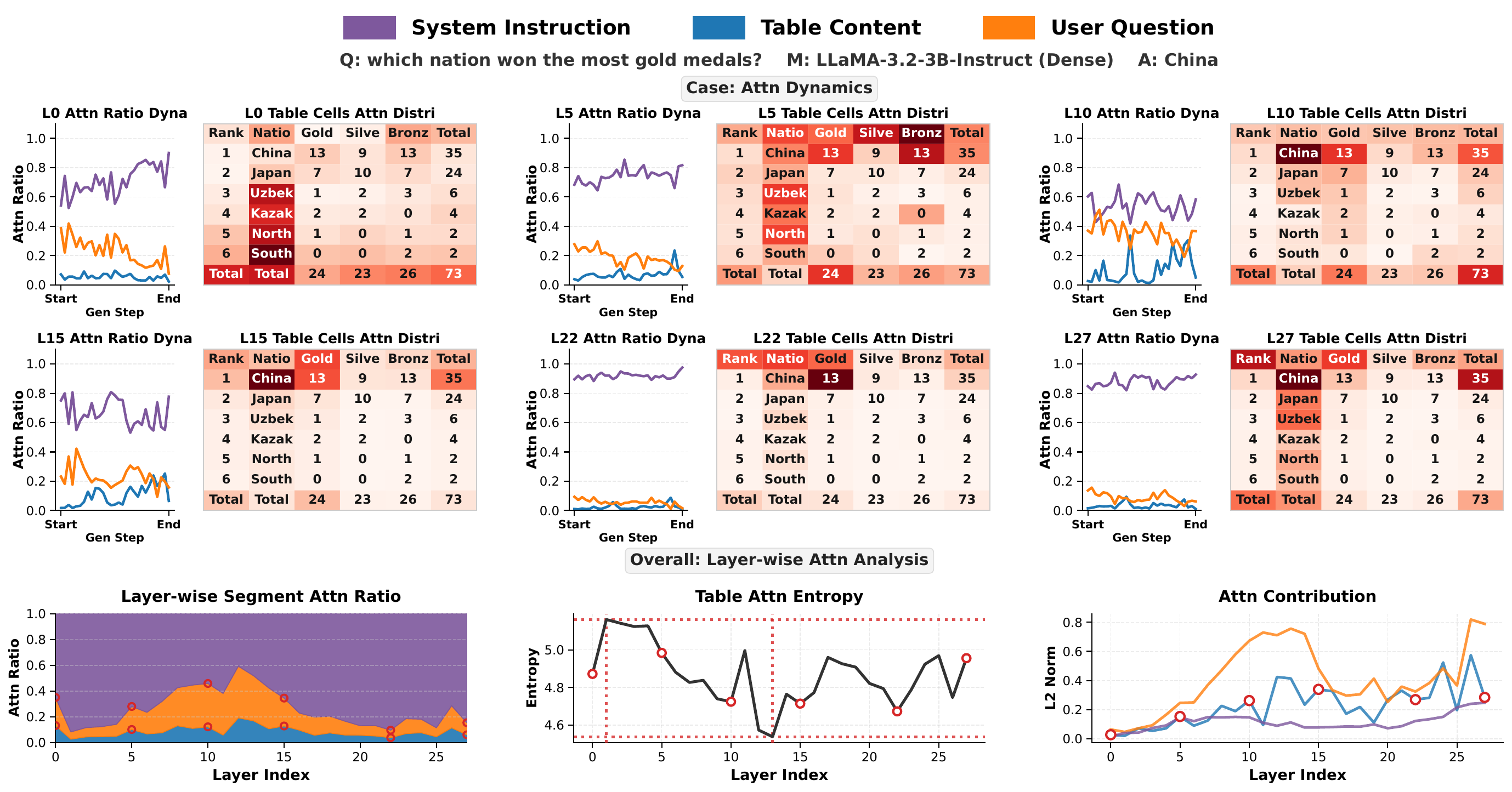}
  \caption{Attention dynamics of LLaMA3.2-3B-Instruct.}
  \label{llama3_2_3b_main_analysis}
\end{figure*}

\begin{figure*}[t]
  \centering
  \includegraphics[width=\linewidth]{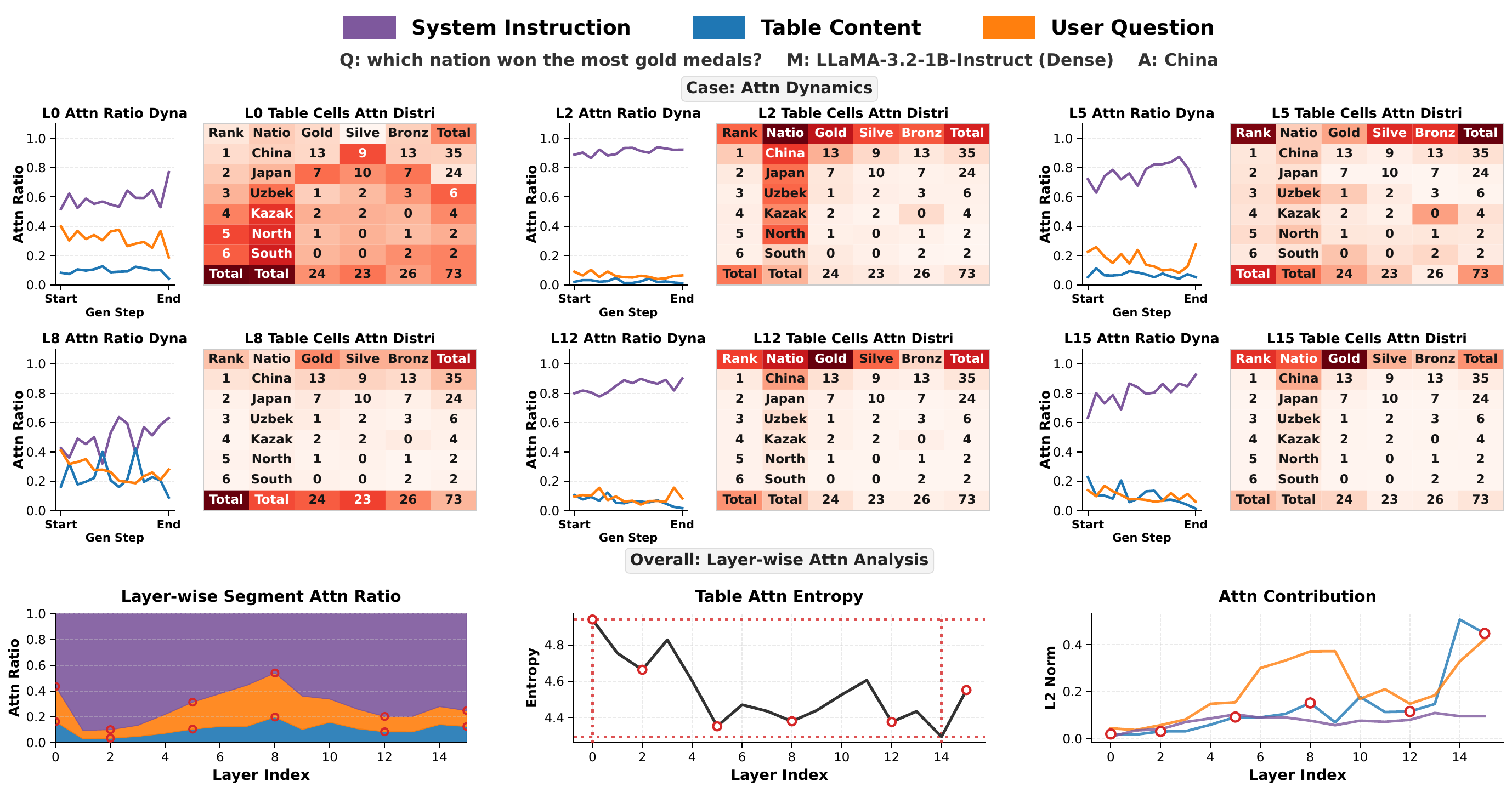}
  \caption{Attention dynamics of LLaMA3.2-1B-Instruct.}
  \label{llama3_2_1b_main_analysis}
\end{figure*}

\begin{figure*}[t]
  \centering
  \includegraphics[width=\linewidth]{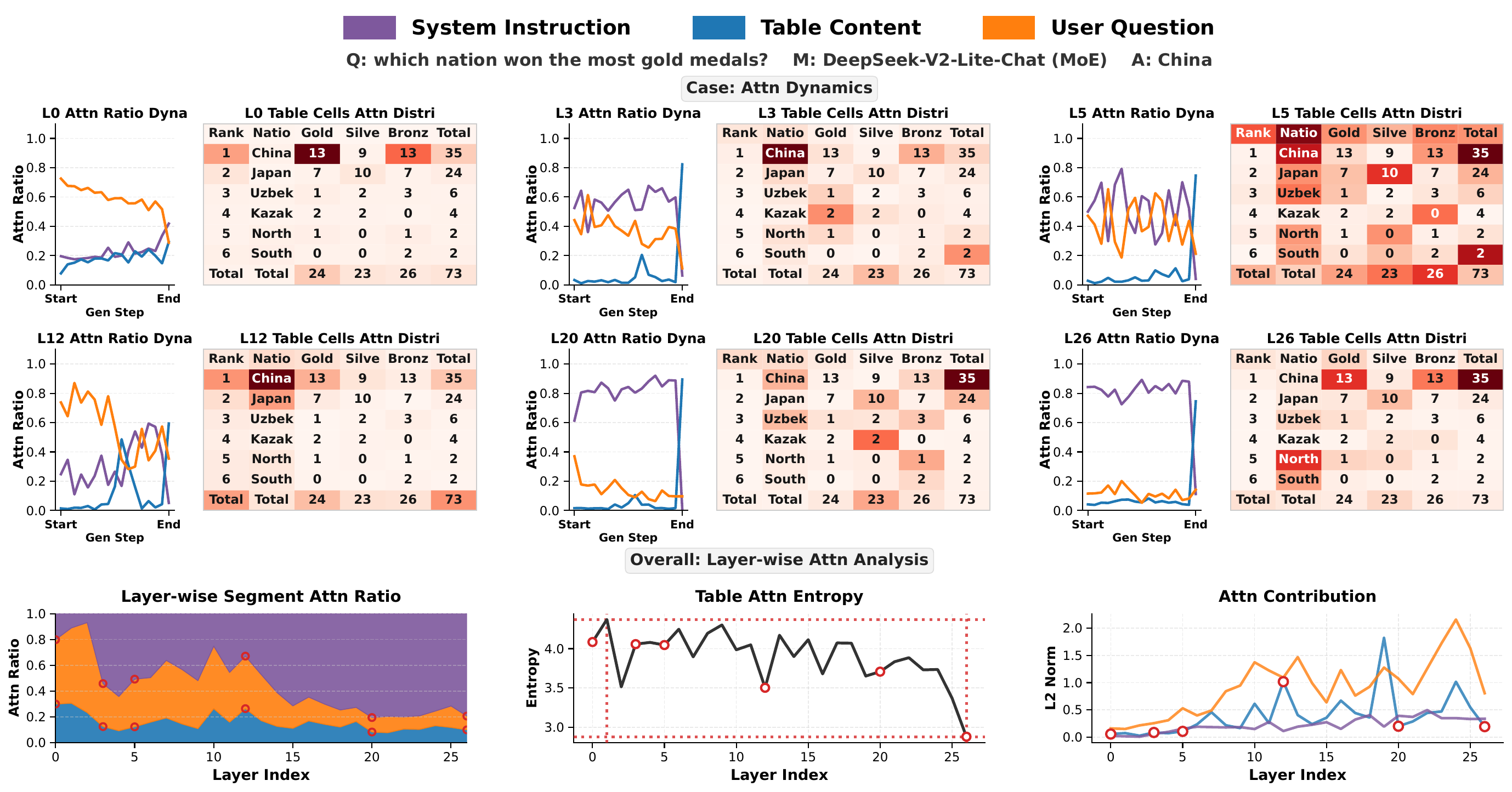}
  \caption{Attention dynamics of DeepSeek-V2-Lite-Chat.}
  \label{deepseek_v2_lite_chat_main_analysis}
\end{figure*}

\begin{figure*}[t]
  \centering
  \includegraphics[width=\linewidth]{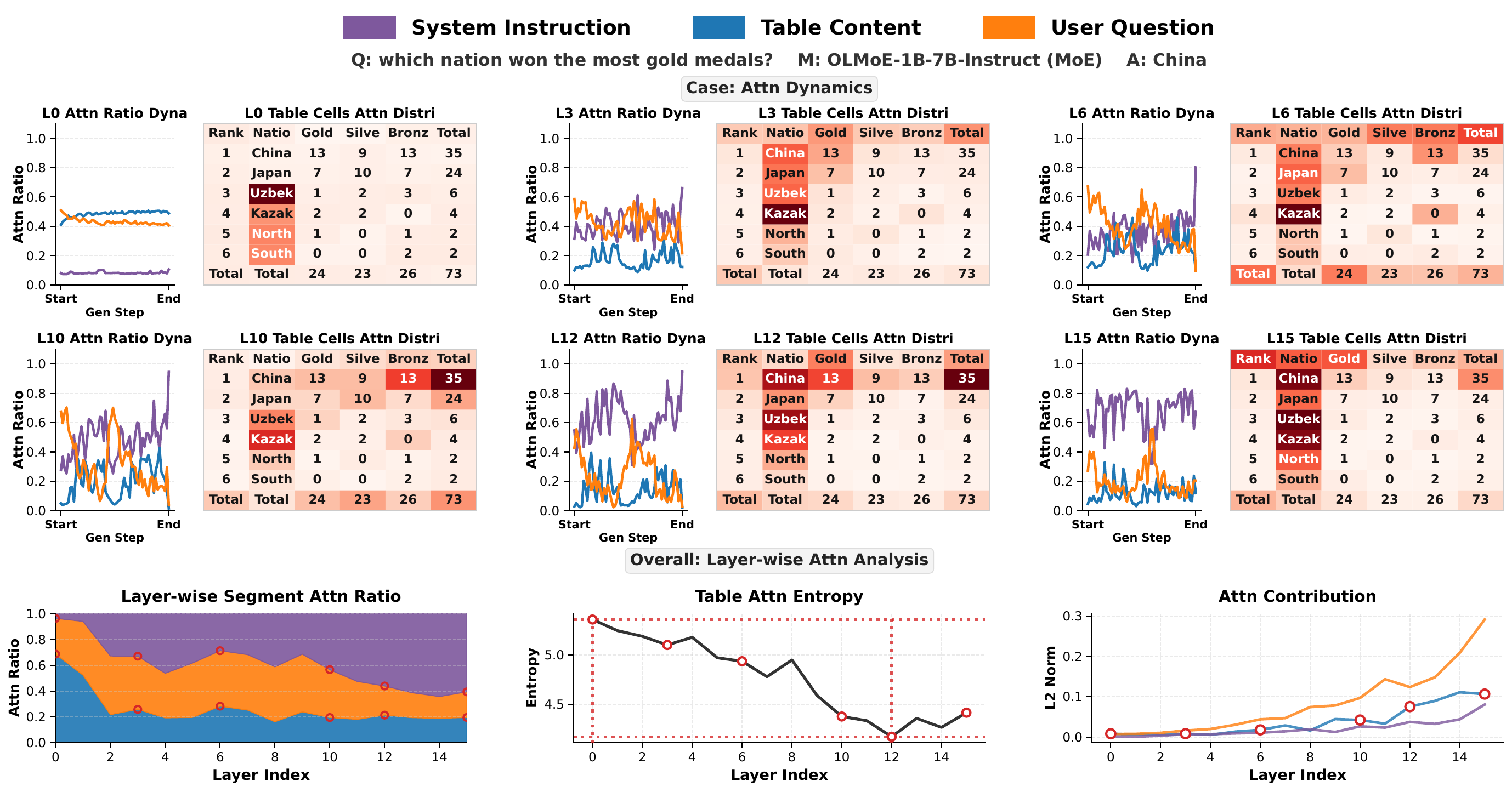}
  \caption{Attention dynamics of OLMoE-1B-7B-Instruct.}
  \label{olmoe_1b_7b_ins_main_analysis}
\end{figure*}

\begin{figure*}[t]
  \centering
  \includegraphics[width=\linewidth]{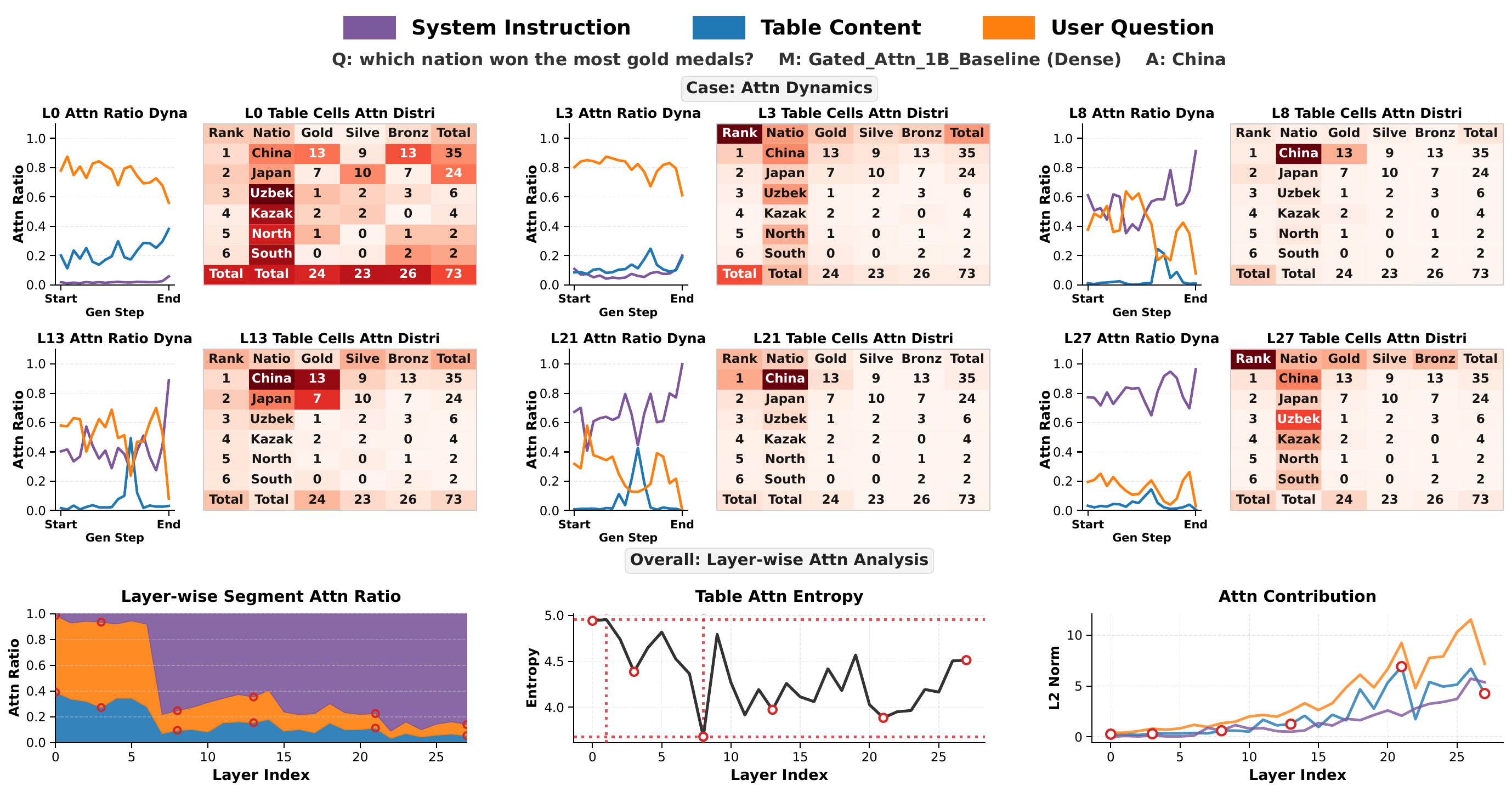}
  \caption{Attention dynamics of Qwen's 1B gated attention baseline.}
  \label{gated_attn_baseline_main_analysis}
\end{figure*}

\begin{figure*}[t]
  \centering
  \includegraphics[width=\linewidth]{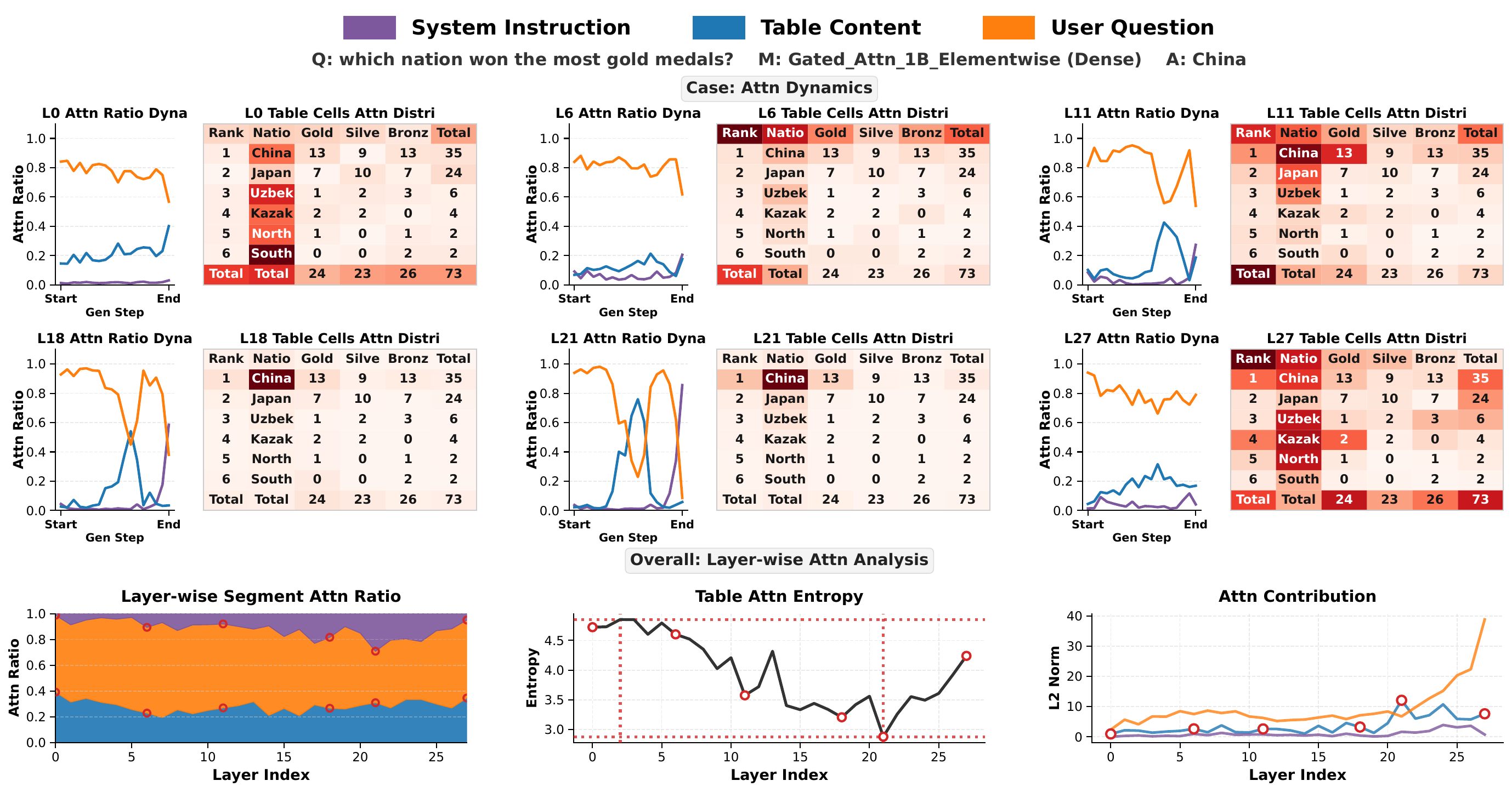}
  \caption{Attention dynamics of Qwen's 1B model with element-wise gated attention.}
  \label{gated_attn_elementwise_main_analysis}
\end{figure*}

\begin{figure*}[t]
  \centering
  \includegraphics[width=\linewidth]{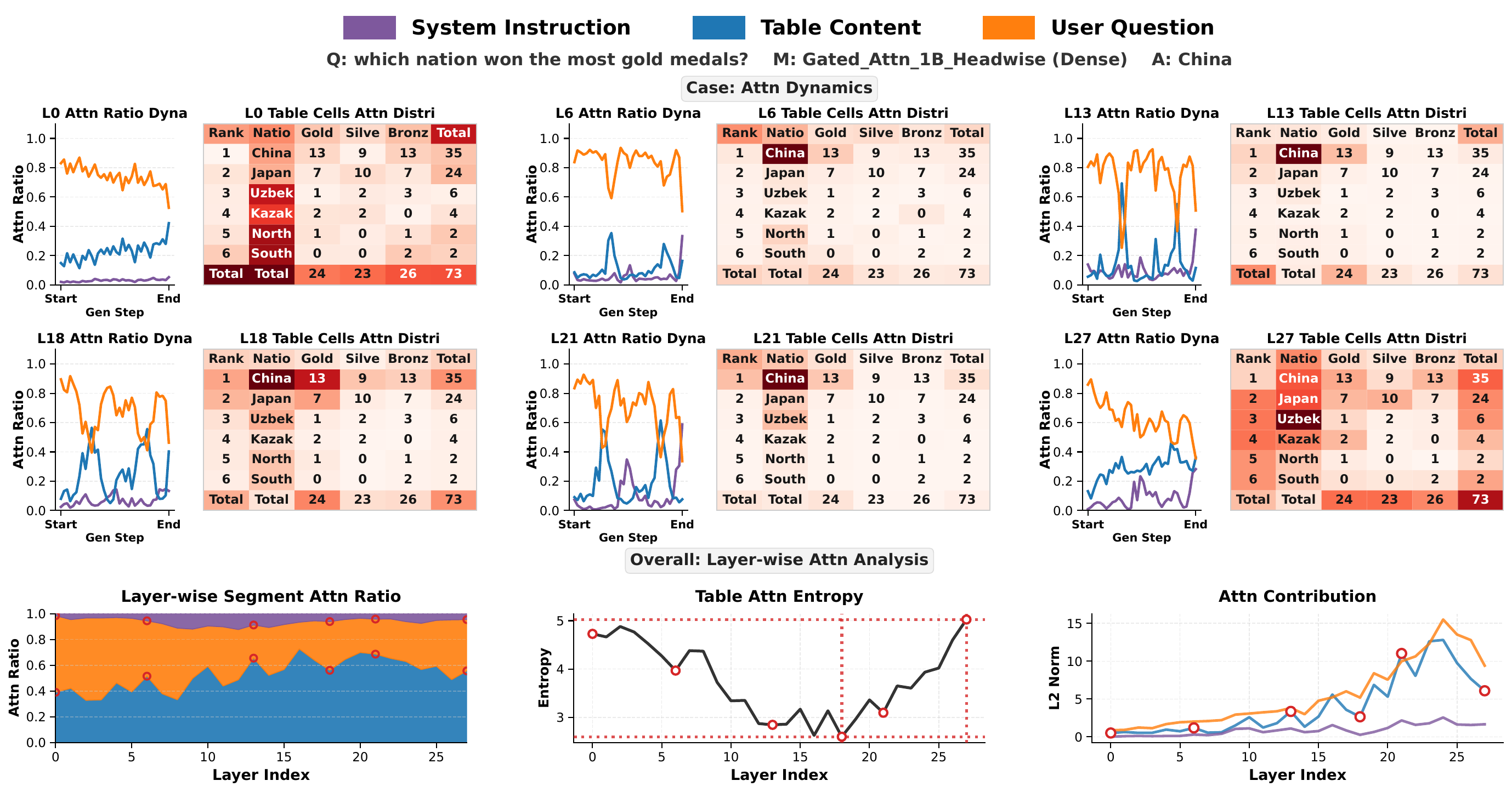}
  \caption{Attention dynamics of Qwen's 1B model with Headwise Gated Attention.}
  \label{gated_attn_headwise_main_analysis}
\end{figure*}


\end{document}